\documentclass[journal,onecolumn]{IEEEtran}
\usepackage{todonotes}
\usepackage{soul}
\usepackage{graphicx}
\usepackage{subfigure}
\usepackage{adjustbox}
\usepackage{caption}
\usepackage{amsmath}
\usepackage{amssymb}
\usepackage{xcolor}
\input{def1b.set}
\usepackage{amsthm,amsfonts}
\usepackage{url}
\usepackage{hyperref}
\usepackage{calrsfs}
\usepackage{algorithm2e}
\DeclareMathAlphabet{\mathcal}{OMS}{cmsy}{m}{n}
\usepackage{stmaryrd}
\usepackage{algorithmicx}
\usepackage{setspace}
\usepackage{multirow}
\usepackage[noend]{algpseudocode}
\usepackage{ifthen}
\usepackage{dutchcal}
\usepackage{eucal}

\usepackage{verbatim}
\theoremstyle{definition}%
\newtheorem{defn}{Definition}%
\newtheorem{exa}{Example}

\usepackage{booktabs}
\graphicspath{{images/}}
\usepackage{array}
\newcolumntype{P}[1]{>{\centering\arraybackslash}p{#1}}
\newcolumntype{M}[1]{>{\centering\arraybackslash}m{#1}}

\ifCLASSINFOpdf
\else

\fi

\begin{document}

\title{Image Reconstruction using Superpixel Clustering and Tensor Completion}
\author{
Maame G. Asante-Mensah$^{*\dag}$,~Anh~Huy~Phan$^{\dag}$,~\IEEEmembership{Member, IEEE},~Salman Ahmadi-Asl$^{\dag}$,~{Zaher Al Aghbari}$^{\ddag}$
 ~and~Andrzej~Cichocki$^{\dag}$,~\IEEEmembership{Fellow, IEEE}

%~Anh~Huy~Phan,$^{\dag}$~\IEEEmembership{Member, IEEE},~ Ivan~Oseledets$^{\dag}$,
%\thanks{This paragraph of the first footnote will contain the date on which you submitted your paper for review. It will also contain support information, including sponsor and financial support acknowledgment. For example, ``This work was supported in part by the U.S. Department of Commerce under Grant BS123456.'' 
%}
\thanks{$^{\dag}$Skolkovo Institute of Science and Technology (SKOLTECH), CDISE, Moscow, $^{*}$Corresponding author E-mail: gyamfua.asantemensah@skoltech.ru.\,\, 
$^{\ddag}$Department of Computer Science, University of Sharjah, Sharjah, 27272, United Arab Emirates.
%Authors were partially supported by the Ministry of Education and Science of the Russian Federation (grant 14.756.31.0001).
}
}
\maketitle
\doublespacing
\begin{abstract}
This paper presents a pixel selection method for compact image representation based on superpixel segmentation and tensor completion. Our method divides the image into several regions that capture important textures or semantics and selects a representative pixel from each region to store. We experiment with different criteria for choosing the representative pixel and find that the centroid pixel performs the best. We also propose two smooth tensor completion algorithms that can effectively reconstruct different types of images from the selected pixels. Our experiments show that our superpixel-based method achieves better results than uniform sampling for various missing ratios.

% This paper proposes a methodology to sample important pixels of images instead of random sampling for image compact
% representation. The proposed approach consists of two stages: 1-smart pixel selection and 2- tensor completion. In the first stage,
% a part of the pixels of an image is selected while in the second stage, we reconstruct the image using efficient tensor completion
% algorithms. Contrary to current works which select the pixels randomly, we proposed to apply the superpixel strategy by first
% dividing the image into several partitions each of which contains some important textures/semantics of the images. Then, from
% each superpixel, we select a representative pixel to be stored. This selection procedure is smart in the sense that important pixels
% are selected from an image. Different methods have been used to sample a pixel from each superpixel but our experimental results
% show the best performance for the centroid pixels. Besides, we developed two smooth tensor completions algorithms which work
% well for reconstructing different types of image categories. Results from simulations and experiments show that our superpixel
% approach outperforms the uniform sampling in all the experiments with different missing ratios.
\end{abstract}


\IEEEpeerreviewmaketitle

\section{Introduction}
Dealing with incomplete data tensors is inevitable in real-world applications due to sensor malfunctioning, inaccurate data acquisition, communication problems or inappropriate handling. However, data elements can also be manually removed to optimise space requirements or to remove unwanted outliers. Estimating the unknown elements of an incomplete data tensor is known as {\it tensor completion} and has played important roles in many machine learning problems, e.g. image/video completion \cite{song2019tensor} and recommender systems \cite{frolov2017tensor}. Due to the importance of tensor completion in the above-mentioned applications, several efficient algorithms have been developed during the past few decades to solve it. Indeed, tensor completion has its basis from matrix completion \cite{candes2009exact}. Similar to the matrix completion where the main assumption is low-rank property of the underlying data matrix, for the tensor case, it is also assumed that the data tensor has low-rank structure. This assumption plays a key role in the formulation of the optimization problem and also deriving theoretical results. It should be noted that the notion of rank for tensors is not unique as it is for matrices, different types of tensor ranks have been defined \cite{cichocki2016low}. Minimizing the matrix rank is an NP hard problem and it has been proved that the nuclear norm is the convex envelope of the matrix rank which can be used to efficiently estimate the matrix rank \cite{fazel2003matrix}. Replacing the matrix rank function with the nuclear norm, converts a non-convex optimization problem into a convex one which is more favorable. Following the matrix case, the nuclear norm of the unfolding matrices has been used to efficiently recover different types of tensor ranks, see \cite{song2019tensor} for a comprehensive review on this topic.

This paper is inspired by the idea of using tensor completion technology to compress and transmit images \cite{li2010tensor}. The idea is to sample a subset of pixels from an image at the source and send them to the destination through a network. Then, at the destination, the tensor completion algorithms are applied to the incomplete image to recover it. This can save memory and speed up data transmission. Most existing papers use uniform sampling to select pixels, but we wonder if other heuristic approaches can perform better. To the best of our knowledge, this question has not been investigated before. We propose to cluster a given image into several partitions that have similar characteristics and then select pixels from each partition. As an efficient clustering method, superpixel algorithms have been used in several applications such as image segmentation \cite{albayrak2019automatic, meinhold2018robust,chen2017linear,li2015superpixel} and object detection \cite{yan2015object}. Among several superpixel methods, Simple Linear Iterative
Clustering (SLIC) algorithm \cite{achanta2010slic} is known for its high performance and fast execution time. Due to these key issues, we use it in our work as a preprocessing stage to find partitions with similar texture/semantic. Then, we select pixels from each superpixel separately. Since each superpixel has homogeneous features and share similar pixel information, the preprocessing stage can hep us to avoid sampling many redundant and similar pixels. Of course, there are several ways to select pixels from each superpixel such as the pixel located in the center or those are on the border. It is also possible to apply the clustering algorithm repeatedly to each superpixel to further explore important pixels. We have extensively investigated the performance of such different sampling strategies. Our simulation results confirmed that the pixel selection based on superpixel preprocessing provides better results than the uniform sampling for a variety of images. Besides, the superpixel preprocessing method with center pixel selection achieved the best results compared with selecting other pixels in the superpixels.  

For recovering the sampled/incomplete image, we also propose the Smooth Tensor nuclear norm (STNN) and Smooth Matrix Nuclear Norm (SMNN) algorithms for image completion. In particular, the smoothing processing enables the completion algorithms to provide better results and this is shown experimentally in our simulation results. The important pixel selection followed by applying efficient tensor completion algorithms totally improves the traditional uniform sampling approach.   

We summarize our main contributions as follows:
\begin{itemize}
    \item We sample and capture important pixels in images using the superpixel technique.
    
    \item We develop efficient tensor completion algorithms with smoothing and filtering methods to enhance their performance.
    
    \item We conduct extensive simulations on various images with missing patterns, including {\it random } and {\it structured} ones.
\end{itemize}

% \textcolor{black}{After this brief introduction, we
% present the notations and concepts that we need in developing our approach in Section \ref{Pre:Sec}}. Section \ref{Sec:Super}, discusses the superpixel technique and its mechanism for extracting important content/semantic of the images which will be used in a part of our approach. In Section \ref{Method:Sec} we study the importance of pixel selection and how to sample important pixels. Section \ref{Sec:Tencom} is devoted to developing tensor completion algorithms where an efficient filtering/smoothing technique is used to improve their performance. Extensive experiments are conducted in Section \ref{Experiments:Sec} to show the applicability and feasibility of the proposed approaches. Finally a conclusion is given in Section \ref{Conclude:Sec}. 

The rest of this paper is organized as follows: Section \ref{Pre:Sec} introduces the notations and concepts that we use in our approach. Section \ref{Sec:Super} explains the superpixel technique and how it extracts important content or semantics from the images. Section \ref{Method:Sec} studies the importance of pixel selection and how to sample important pixels. Section \ref{Sec:Tencom} develops tensor completion algorithms with an efficient smoothing technique to improve their performance. Section \ref{Experiments:Sec} presents extensive experiments to demonstrate the applicability and feasibility of our methods. Section \ref{Conclude:Sec} concludes the paper.

%however, there has been a lot of studies for compression and the method has been proven to work quite well in the area of image compression \cite{ertem2020superpixel,hariharan2018light,rizkallah2018graph,luo2017image,fu2016adaptive,fracastoro2015superpixel}. 

%Because of the limits of the human visual system, if an image is modified within the Just Noticeable Distortion (JND) threshold, it cannot be noticed by the human eye \cite{hariharan2018light}.
%In recent times the approach has also been applied to neural networks however the computational complexity and the..
%Superpixel segmentation with fully convolutional networks\cite{yang2020superpixel}. 
%In this work, we simultaneously exploit the globally multidimensional structure and locally piecewise smoothness to further enhance the performance of the method.

%\end{comment}
\section{Notations and definitions}\label{Pre:Sec}
%\todo[inline]{We need to rephrase this section}

Basic notations used in this work  are taken from \cite{cichocki2016low}. We represent scalars by the lowercase letters. A vector is given by a boldface lower case letter, e.g. $\ba$. A matrix is represented by boldface capital letter, e.g. $\bA$ and a higher order tensors are also denoted by bold underlined capital letter, e.g. $\underline{\bA}$. 
%Also, matrices and tensors  are graphically represented as shown in Figure \ref{fig:Graphrepres} (a).
%The $I_n$ designate the number of elements in the respective dimension. 
%The $(i_1,i_2,\ldots, i_N)$th element of an $N$th-order tensor $\underline{\bA} \in \Real^{I_1 \times I_2 \times \cdots \times I_N }$ is denoted by $x_{i_1,i_2,\ldots, i_N} = \underline{\bX}({i_1,i_2,\ldots, i_N}) $. 
For an $N$th-order tensor $\underline{\bA} \in \Real^{I_1 \times I_2 \times \cdots \times I_N }$ its $(i_1,i_2,\ldots, i_N)$th element is denoted by $x_{i_1,i_2,\ldots, i_N}$ and is represented as $\underline{\bA}({i_1,i_2,\ldots, i_N}) $.
%Trace and Moore-Penrose pseudoinverse of matrices are represented by “Tr” and $\dag$, respectively.
%The order of a tensor can also be referred to as its “modes”,  “ways” or “dimensions”. This represent the  space,  time, frequency,  trials,  classes,  and  dictionaries of the time series data.    
%The columns of the matrix $\bA= [\ba_1,\ba_2,\ldots, \ba_J] \in \Real ^{I\times J} $ are the vectors, where each elements of a matrix are the scalars given as, $a_{ir}=\bA(i,r)$. 
%The set $\{\underline{\bX}^{(n)}\}^N_{n=1} :=
%{\underline{\bX}^{(1)} , \underline{\bX}^{(2)}, \ldots, %\underline{\bX}^{(N)} }$ represents a tensor sequence, with %$\underline{\bX}^{(n)}$  being the $n$th-order tensor of the %sequence. The representations of matrix sequences and vector %sequences are designated in the same way. 
%The multi-index is defined as $\overline{{i_1}{i_2} \ldots {i_N}} = {i_N} + \left( {{i_{N - 1}} - 1} \right){I_N} +  \cdots  + \left( {{i_1} - 1} \right){I_2}{I_3} \cdots {I_N}$.

The $n$-mode \textit{matricization} of a tensor $\underline{\bA} \in \Real^{I_1 \times I_2 \times \cdots \times I_N }$ which is also called mode-$n$ unfolding of a tensor \cite{kolda2009tensor} is shown by $\bA_{(n)}\in \Real^{ I_n \times I_1 \cdots I_{n-1}I_{n+1} \cdots I_N }$.  \textit{Tensorization} or matrix folding is the process of converting a low-order tensor to a higher-order tensor. When we fix all indices except two of a tensor, a sub-tensor is generated and it is called a slice. For example, for a tensor $\underline{\bf A}\in\mathbb{R}^{I \times J\times K}$, the slices $\underline{\bf A}(:,:,k),\,i=1,2,\ldots,I_3,$ are called frontal slices and is denoted as ${\bf A}_{(k)}$,  the tube in a tensor is denoted as $\underline{\bf A}(i,j,:)$.
%We define the folding operation for the first type of mode-$n$ unfolding as ${\rm fold}_n(\boldsymbol{\cdot} )$, e.g. a tensor $\underline{\bX}$, we have its folding operation as $ {\rm fold}_n(\bX_{(n)} ) = \underline{\bX}$.
The inner product of two tensors $\underline{\bA}$ , $\underline{\bB}\in\Real^{I_1 \times I_2 \times \cdots \times I_N }$ is defined
as $\langle \underline{\bA},\underline{\bB} \rangle  =  \sum_{i_1} \sum_{i_2} \ldots \sum_{i_N} a_{i_1,\ldots, i_N} b_{i_1,\ldots, i_N}$ and the Frobenius norm of a tensor is given as $\|\underline{\bA}\|_{F}= \sqrt{\langle \underline{\bA},\underline{\bA} \rangle}$.

\subsection{Low-Rank Tensor Completion}\label{TTTR_C:Sec}
Let an incomplete data tensor $\underline {\bA}\in\Real^{I_1 \times I_2 \times \cdots \times I_N }$, be given and assume the indices of its observed elements are arranged in the indexing binary tensor  $\boldsymbol{\underline{\Omega}}\enspace\in\mathbb{R}^{I_1\times I_2\times \cdots\times I_N}$. The complement of the indexing set $\boldsymbol{\underline{\Omega}}$ is represented as ${\underline{\bf \Omega} ^ \bot }(i_1,i_2,\ldots,i_N)$. \textcolor{black}{The projection operator $\boldsymbol{\underline{\Omega}}$ over the data tensor $\underline{\bf A}$ is defined  as follows:}
%using the operator ${{\bf P}_{\underline{\bf \Omega}} } \left( \underline{\bf A} \right) $ 
\[
{\mathcal{P}_{\underline{\bf \Omega}} } \left( \underline{\bf A} \right) = \left\{ \begin{array}{l}
{a_{{i_1},{i_2}, \ldots ,{i_N}}}\,\,\,\,\,\,\left( {{i_1},{i_2}, \ldots ,{i_N}} \right) \in \underline{\bf \Omega}. \\
0\,\,\,\,\,\,\,\,\,\,\,\,\,\,\,\,\,\,\,\,\,\,\,\,\,\,\,\,\,\,\,{\rm for \, missing \, entries}.
\end{array} \right .
\]
The task of low rank tensor completion can be formulated as the following optimization problem
% \textcolor{black}{and finally the optimization problem is presented as:}
\begin{equation}\label{MinProb}
\arg \min_{{\underline{\bf B}}} \,\,\left\| { {\mathcal{P}_{\underline{\bf \Omega}} } \left( \underline{\bf B} \right) - {\mathcal{P}_{\underline{\bf \Omega}}} \left({\underline{\bf A}} \right) } \right\|^{2}_F \, ,
\end{equation}
where $\underline{\bf B}$ is the estimated data tensor. To make the problem \eqref{MinProb} well-posed we need to impose constraints on the data tensor $\underline{\bf B}$, e.g. low-rank property. Depending on the tensor rank notion defined such as Tucker rank, Tensor Tran rank, Tubal rank etc, the minimization problem \eqref{MinProb} is solved over the space of tensors with at most the predefined tensor rank. If estimating the tensor rank is difficult, then the minimization problem \eqref{MinProb} can be converted to the tensor rank minimization problem formulated as follows
\begin{equation}\label{MinProbtran}
\arg \min_{{\underline{\bf B}}} \,\,{\rm rank} (\underline{\bf B})+\left\| { {\mathcal{P}_{\underline{\bf \Omega}} } \left( \underline{\bf B} \right) - {\mathcal{P}_{\underline{\bf \Omega}}} \left({\underline{\bf A}} \right) } \right\|^{2}_F,
\end{equation}
where by solving problem \eqref{MinProbtran}, we can reconstruct the data tensor and also estimate the tensor rank. Unfortunately, the tensor rank minimization is an NP hard problem and a surrogate or a relaxation of it should be considered (to be discussed in Section \ref{Sec:Tencom}). 
% \[
% \underline{\bf \Omega}(i_1,i_2,\ldots,i_N)= \left\{ \begin{array}{l}
% 1\,\,\,\,{\rm if}\,{a_{{i_1},{i_2}, \ldots ,{i_N}}}\,{\rm is\,known,}\\
% 0\,\,\,{\rm if}\,{a_{{i_1},{i_2}, \ldots ,{i_N}}}\,{\rm is\,unknown,}
% \end{array} \right.
% \]
% where the complement is written as:
% \[
% {\underline{\bf \Omega} ^ \bot }(i_1,i_2,\ldots,i_N) = \left\{ \begin{array}{l}
% 0\,\,\,\,{\rm if}\,{a_{{i_1},{i_2}, \ldots ,{i_N}}}\,{\rm is\,known.}\\
% 1\,\,\,\,{\rm if}\,{a_{{i_1},{i_2}, \ldots ,{i_N}}}\,{\rm is\,unknown.}
% \end{array} \right.
% \]
%The operator ${{\bf P}_{\underline{\bf \Omega}} } \left( \underline{\bf Y} \right) $ projecting the data tensor $\underline{\bf Y}$ onto the observation index tensor $\boldsymbol{\underline{\Omega}}$ is defined as
%We will utilize the idea of superpixel approach to sample important pixels in the images. For this reason, in the next section, we describe this technology and explain how it works.

\section{Superpixel clustering}\label{Sec:Super}
A group of pixels that share similar features such as pixel intensity, texture and color are referred to as superpixels. Superpixels can be found in many applications of computer vision and machine learning tasks.
As an efficient superpixel method, SLIC (Simple Linear Iterative Clustering) algorithm was introduced in \cite{achanta2010slic} and due to its superior performance \cite{malisiewicz2007improving,neubert2012superpixel} was extensively used in many applications such as image segmentation \cite{albayrak2019automatic, meinhold2018robust,chen2017linear,li2015superpixel}, object detection \cite{yan2015object}, anomaly detection \cite{zhou2021proxy,sakurada2015change} and image reconstruction \cite{cao2021dynamic,kumar2019superpixel,zitnick2007stereo}.
%However the limitation of this method is the computational effort needed and the risk of loosing meaningful image information \cite{neubert2012superpixel}. As a result, careful choices of the superpixel algorithm parameters for the particular application is very important for the success of the clustering. 
%First we apply the super-pixel clustering which acts as the initial segmentation stage. Here image pixels are not removed uniformly but instead, our clustering method apply SLIC (Simple linear iterative clustering) method in Matlab to group pixels into regions with similar values.
%The SLIC algorithm is a k-means-based clustering of
%neighboring pixels by considering their color and coordinate information.
%The SLIC algorithm adapts a k-means clustering approachto efficiently generate superpixels. 
\textcolor{black}{To efficiently generate superpixels, the SLIC algorithm employs a k-means clustering approach.}
It basically converts an RGB color space into the CIELAB color space as $[l,\, a, \,b,\, x, \, y]^T $ where $[l, \, a, \, b]^T$ is the pixel color vector in CIELAB color space and $[x, \, y]^T$ is the pixel position.
The spatial distances are normalized in order to use the Euclidean distance in the 5D space. To cluster pixels in 5D space, a new distance measure that takes super-pixel size into account is introduced. 
This method takes as input a desired number of approximately equally sized superpixels $K$. At first, $K$ superpixel cluster centers $\bc_{k},\,k=1,2,\ldots,K$ are chosen at regular grid intervals $S$. 
Since the spatial vastness of any superpixel is approximately $S^2$, we assume that pixels associated with this cluster center are located within a $2S \times 2S$ area on the xy plane surrounding the superpixel center. $D_{s}$ is the normalized distance measure that will be used in 5D space and defined as follows
\begin{equation}\label{superpixeldistance}
\begin{array}{l}
D_{s} = d_{lab} + (m/S)\ast  d_{xy},\\
d_{lab}=\sqrt{(l_k - l_i)+(a_k - a_i)+(b_k - b_i)},\\
d_{xy}= \sqrt{((x_k - x_i)^{2} + (y_k - y_i)^{2}) },
\end{array}
\end{equation}
where $i$ represents the value to be clustered. The sum of the lab distance $d_{lab} $ and the xy plane distance $d_{xy}$ normalized by the grid interval $S$. Besides, the distance measure $D_{s}$ includes a variable $m$ that allows us to control the compactness of a superpixel.
The greater the value of $m$, the cluster becomes more compact. This value can range between 1 and 20. The SLIC algorithm is summarized in the Appendix (Algorithm \ref{ALG:SLIC}). See  Figure \ref{fig:house_superpixel} for a graphical illustration on the SLIC algorithm for $K=50,100,200$ superpixels.

%\section{Related works on superpixel segmentation and clustering}
\textcolor{black}{The use of superpixels for various computer vision and image processing tasks acts as a preprocessing step to reduce the complexity of subsequent processing. They are also used to capture redundancy in an image \cite{ neubert2012superpixel}.
A superpixel clustering method was employed in \cite{baya2020pixel} as an initial step to segmentation, texture learning and patch matching for image reconstruction. Indeed, the superpixel methods as clustering techniques allow for sampling important pixels used in the reconstruction of the images. The reconstructed images can then be used to perform other tasks achieving results comparable to the original image.
The clustering techniques using superpixel have been implemented in many neural networks and autoencoders models. These methods have proven to be very effective with promising results. The unsupervised learning models are very useful for feature extraction. A  Dual Graph Autoencoder (DGAE), proposed by Zhang et al. \cite{ zhang2022spectral} constructs the superpixel-based similarity graph using entropy rate superpixel which captures the spatial information in the image and generate a band-based similarity graph that can be used to characterize the geometric structures of hyperspectral images. The dual graph convolution, allows more discriminative feature representations to be learned from the hidden layers that aids in the generation of a clustering map. Superpixels has also seen implementation in medical data segmentation \cite{tian2015superpixel,bechar2018semi,ouyang2020self,huang2021dense}. The implementation by Bechar  et al \cite{ bechar2018semi} uses a semi-supervised model for optic cup and disc segmentation to calculate the cup to disc ratio value.  Here, the SLIC superpixel was adopted for generating some labels used in the retina segmentation. However, the drawback to most of these sophisticated models is hyperparameter fine-tuning and also the computational load needed to achieve the desired result. Superpixel based image representation in \cite{tasli2015superpixel} acquires mid-level information in order to improve the object recognition accuracy. Even though the main work focuses on the image recognition task, superpixel clustering has been performed specifically for the feature extraction step.}

% \RestyleAlgo{ruled}
% \begin{algorithm}
% \LinesNumbered
% \SetKwInOut{Input}{Input}
% \SetKwInOut{Output}{Output}
% %\Input{Initialize cluster centers $c_k = [l_k,a_k,b_k,x_k,y_k]^{T}$ by sampling pixels at regular grid steps $ S$}
% %\Output{Completed data tensor ${\underline{\bf X}}$}
% \caption{The SLIC method \cite{achanta2010slic}} \label{ALG:SLIC}
%       \textbf{Set} clusters centers $c_k = [l_k, \, a_k, \, b_k, \, x_k, \, y_k]^{T}$ by taking regular grid steps $S$ to sample pixels \\
%       \textbf{shift} Cluster centers in an $ n\times n$ neighborhood  to the lowest gradient location\\

% \While{$\epsilon \geq$ threshold }
% {
%       \For{each cluster center $c_k$}
%       {
%       \textbf{Assign} the best matching pixels from a $2S \times 2S$ square  neighborhood around the cluster center $c_k$ using the distance measure for $D_s$  in equation \ref{superpixeldistance}
%       }
%     \textbf{Compute} cluster centers \\
%     \textbf{Calculate}  the residual error $\epsilon$ using the $L_1$ distance between recomputed centers and previous centers. 
% }
% \end{algorithm}

\begin{figure*}[ht!]
    \centering
    \includegraphics[width=0.6\linewidth]{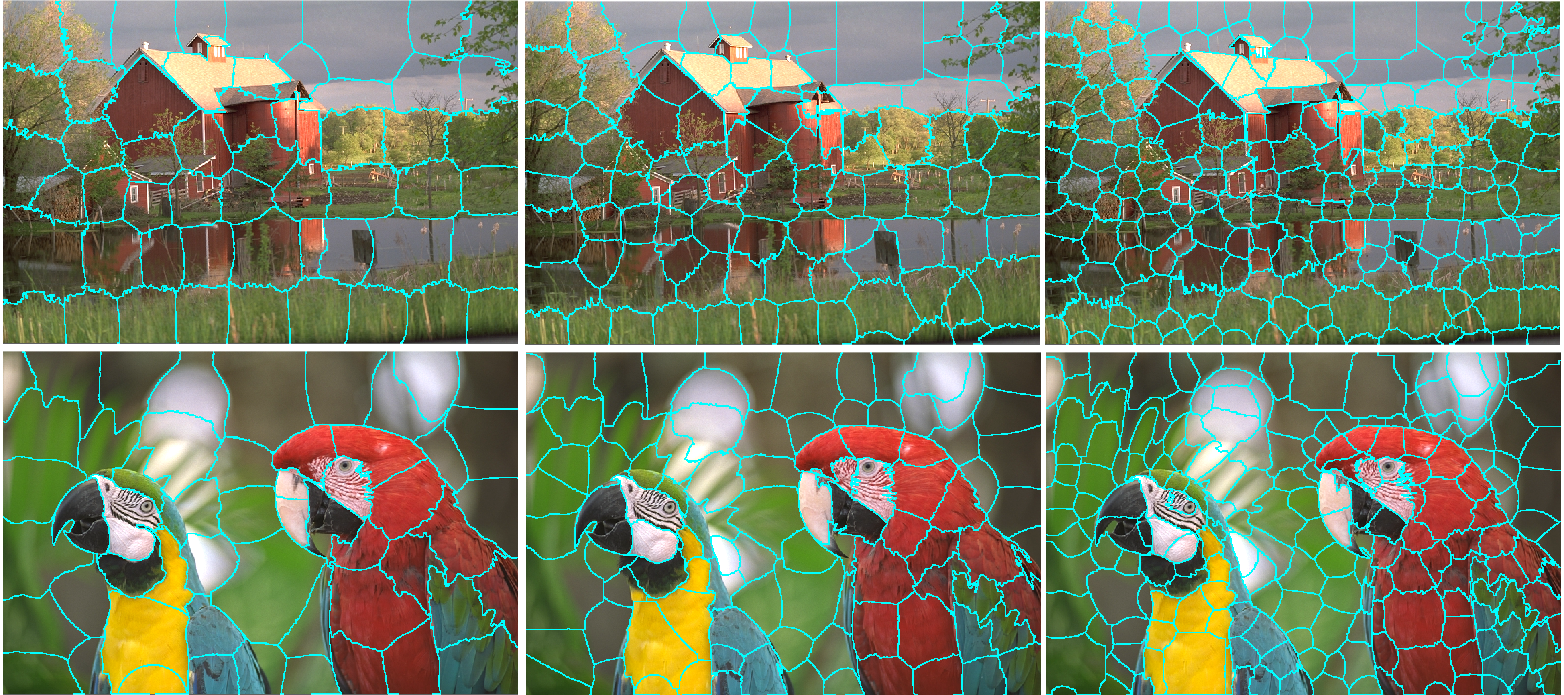} 
    \caption{Superpixel clustering using SLIC with $k = 50, 100,$ and $200$ superpixels.}
    \label{fig:house_superpixel}
\end{figure*}

\begin{figure*}[ht!]
    \centering
    \includegraphics[width=0.7\linewidth]{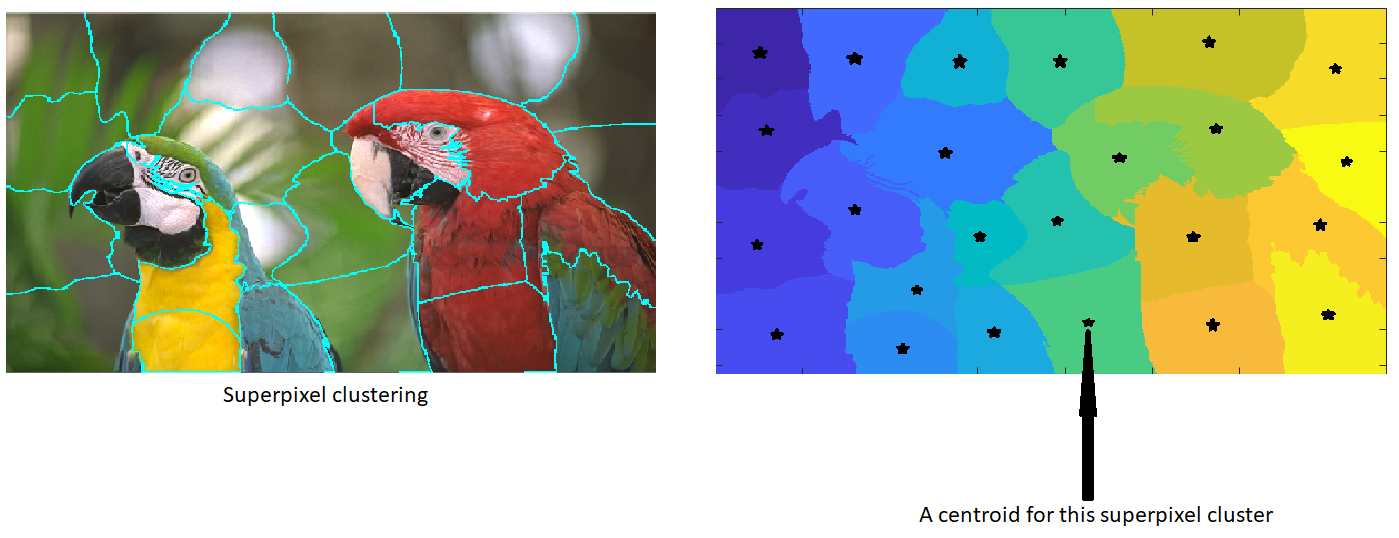} 
    \caption{Illustration of centroid superpixel sampling.} 
    \label{fig:centroid}
\end{figure*}

\begin{figure*}[ht!]
    \centering
    \includegraphics[width=0.7\linewidth]{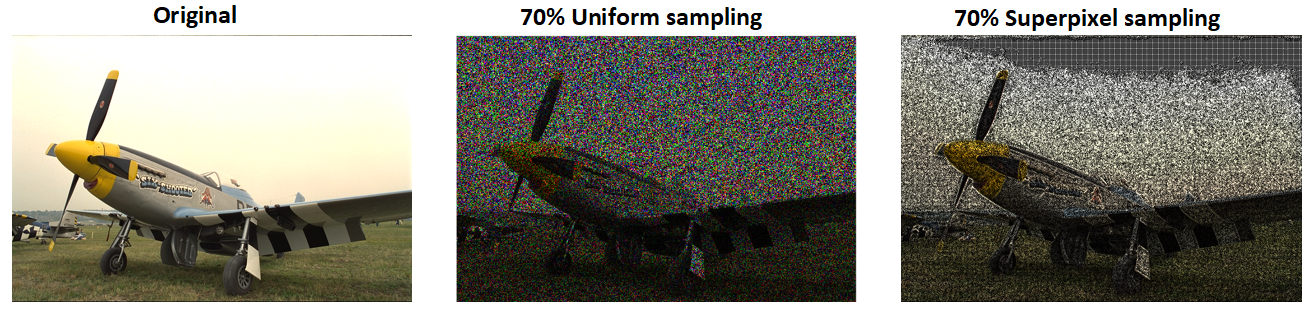} 
    \caption{Illustration of uniform sampling vs superpixel sampling on kodim20 image. Uniform sampling can not find the important pixels and treat them accordingly while the superpixel approach chooses the most promising pixels.} 
    \label{fig:uniform_centroid}
\end{figure*}

\begin{figure*}[ht!]
    \centering
    \includegraphics[width=0.9\linewidth]{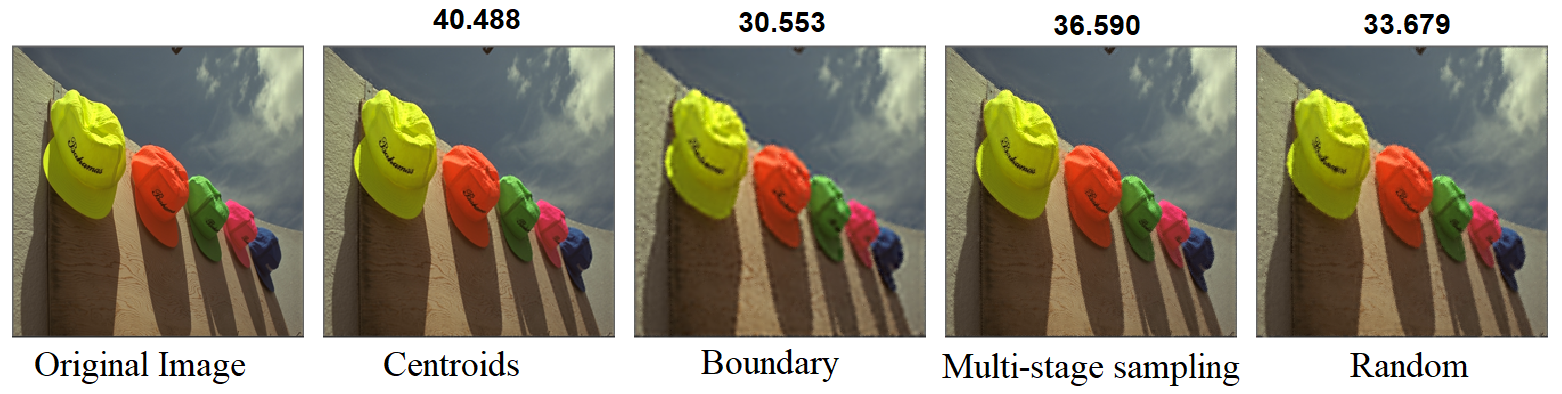} 
    \caption{PSNR comparison of kodim03 image using different superpixels sampling methods with approximately selecting 50\% of the data. The experiment show that centroid pixel selection provides better results.}
    \label{fig:sampling_methods}
\end{figure*}

% \begin{figure*}[ht!]
%     \centering
%     \includegraphics[width=0.52\linewidth]{quantized_illustrationk_all_crop1.png} 
%     \caption{Quantized results of some images using SLIC superpixels clustering } % 1586 clusters
%     \label{fig:quantized}
% \end{figure*}

\section{Proposed Pixel Sampling Method}\label{Method:Sec}

% It is a known fact in signal processing community that if a data tensor admits a low-rank structure, it can be efficiently recovered from a part of its components, see \cite{song2019tensor} for a comprehensive review on this topic. Benefiting from this interesting discovery, many algorithms were developed for the tensor completion problem applied to different types of datasets such as images and videos. There are mainly two types of missing patterns: {\it random} and {\it structured} sampling. The random sampling refers to removing the pixels of images/videos randomly. In the structured sampling, sequential fibers, slices or parts of the data are removed. For example, in \cite{yokota2018missing} slices of videos are removed which a structured missing pattern and Hankelization approach is used to recover the videos.
% Of course the structured missing patterns are more challenging than the random ones as we lose information about of a whole region of that image/video. However, for the case of missing strategy, it has been not well studied the performance of completion algorithms for pixel selection in random or heuristic/smart way. More precisely, given an image and assuming that we want to remove $60\%$ of its pixels, the question is: how to select the best $40\%$ of the pixels so that we can achieve the best image recovery performance. We are interested in investigating this question in this paper.

Signal processing researchers have found that a data tensor with a low-rank structure can be efficiently recovered from a subset of its components \cite{song2019tensor}. This finding has led to many tensor completion algorithms for various types of data, such as images and videos. The missing components can have either random or structured patterns. Random patterns remove pixels of images/videos randomly, while structured patterns remove sequential fibers, slices, or parts of the data. For example, \cite{yokota2018missing} uses a structured pattern to remove slices of videos and recovers them using a Hankelization approach. Structured patterns are more challenging than random ones, because they lose information about a whole region of the data. However, the question of how to select the best subset of pixels for optimal image recovery performance has not been well studied. This paper aims to investigate this question.
Most of the existing papers use uniform sampling to remove pixels of images or videos. For example, in \cite{li2010tensor}, it is proposed to sample some pixels of a given image uniformly and remove the rest to reduce the memory requirement and to transfer it faster in the network. The sampled pixels are recovered in the destination through the tensor completion process. Contrary to the simple pixel selection using random sampling, we propose to first cluster the image to some partitions or so-called superpixels which share common characteristics. Then, we select the pixels from each of the clusters. Since the pixels in each cluster more likely have similar features, the clustering as a preprocessing stage can help to avoid selecting redundant pixels as is done in random sampling. In this sense, it is a kind of smart pixel sampling as we try to select pixels in a heuristic and an intelligent way.  

It is worth pointing out that to select pixels in each cluster, there are several possibilities. For example,  We can select a pixel located in the center of the cluster or on its boundary/border. Depending on a way used to select a pixel from each superpixel, we define the following categories:

\begin{itemize}
\item {\bf Centriod superpixel approach}
The centroid superpixel refers to the case when we select a center of each superpixel to be the selected pixel.

\item {\bf Boundary superpixel approach} This approach selects a pixel from a border/boundry of the superpixel and use it for the sampling procedure.

{\item {\bf Multi-stage superpixel approach.}} It is also possible to cluster each spuerpixel again to further explore important pixels and we call it multi-stage superpixel approach. 
\end{itemize}

\begin{figure*}[ht!]
    \centering
    \includegraphics[width=0.4\linewidth]{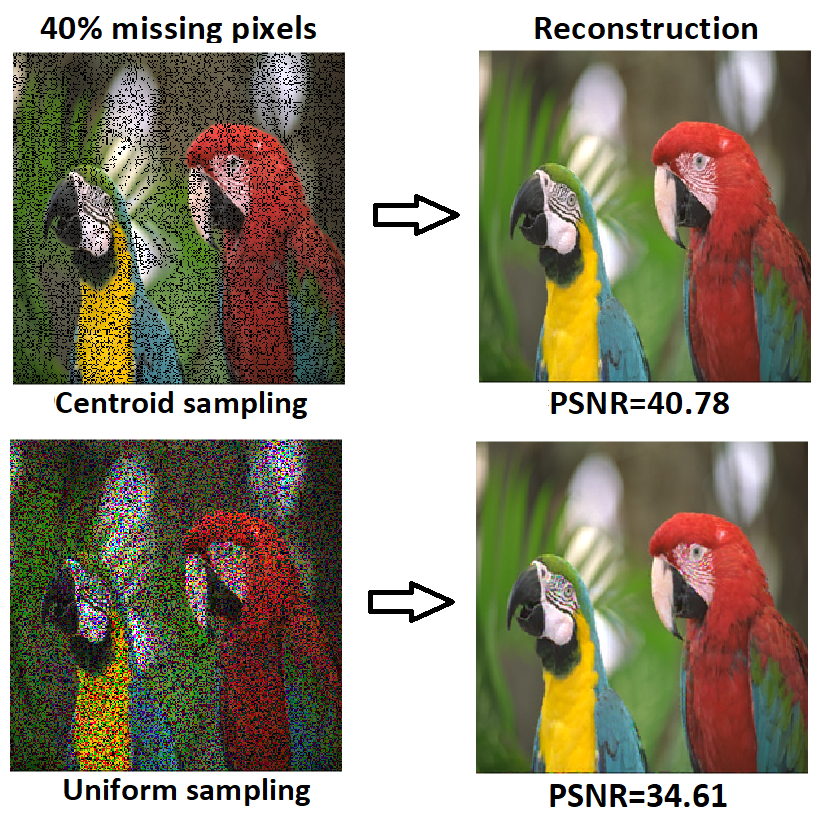} 
    \caption{The results show that centroid pixel selection provides better results than the uniform sampling. For Kodim23 image, we have used 237905 clusters which leads to sampling 60\% of all pixels. 
} 
    \label{fig:centroid_vs uniform}
\end{figure*}

Figure \ref{fig:centroid}, shows the superpixel clustering applied to a given image followed by selecting the center of each superpixel. Also, Figure \ref{fig:uniform_centroid}, demonstrates what looks like an image after sampling $30\%$ of pixels using uniform sampling and superpixel sampling with center selection. We extensively investigated the difference between the performance of different sampling methods for the superpixel approach and compared it with the uniform sampling. Our simulation results show that in most of our experiments on a variety of images, the center of superpixels provides quite promising results. For example, in  Figure \ref{fig:sampling_methods}, the results of applying the superpixel clustering (with center, boundary and multi-stage selection) and the uniform sampling to an image with $50\%$ pixel sampling have been reported. As we earlier mentioned, the superpixel with center pixel selection achieved the best performance. It is interesting to note that for $60\%$ pixel sampling, the difference between uniform sampling and superpixel clustering with center selection was significant as seen in Figure \ref{fig:centroid_vs uniform}. After sampling important pixels, we have a compact variant of the image which can be transmitted and it can be recovered in the destination using tensor completion algorithms (see Section \ref{Sec:Tencom}).

% \begin{figure*}[ht!]
%     \centering
%     \includegraphics[width=0.7\linewidth]{General_Pipleline3_new.png} 
%     \caption{A general pipeline of our proposed reconstruction method}
%     \label{fig:general_pipeline}
% \end{figure*}

\section{Tensor Completion with smoothing/filtering}\label{Sec:Tencom}
As discussed in Section \ref{Method:Sec}, in the first stage of our methodology, we sample a part of images to have a compact representation of the image. This strategy was used in \cite{li2010tensor} to store only a part of pixels. In the second stage, the image with only observed pixels should be recovered based on the completion algorithms. Here, the performance of such completion algorithms is crucial for the better image recovery. In this section, we use the nuclear norm minimization (both matrix and tensor scenarios, see the Appendix) for the completion task. Nonetheless, we apply the smoothing strategy to enhance the performance of the algorithms. In the simulation section, we will show the importance of smoothing strategy for a better image recovery. The matrix completion and tensor completion formulations are presented in Subsections \ref{SEC:MA} and \ref{SEC:TA}, respectively.

\subsection{Tensor completion based on matrix nuclear norm regularization}\label{SEC:MA}  
A significant advantage of the low-rank matrix approximation is that the vital information in a matrix, 
given in terms of degree of freedom, is substantially less than the total number of entries.
As a result, even if the number of observed entries is small, there is still a decent chance of recovering the entire matrix \cite{nguyen2019low}.
This advantage has been shown in computer vision tasks \cite{liu2012tensor,lu2017unified} for estimating missing values in images. In this section, we describe the completion task.
The model of rank minimization based matrix completion is formulated as:
\begin{equation}\label{rank_minimization}
\begin{array}{l}
\mathop {\min }\limits_{\bf X }
\,\,\,\,\,    {{ {\rm rank} \, ({\bf X}) }} \\
%+ \left\| {{{\bf P}_{\Omega}}\left( {\bf X -  {\bf Y }}  \right)} \right\|_F^2\, 
 s.t. \,\,\,\, {\bf X}_{\Omega}={\bf Y}_{\Omega}
\end{array}
\end{equation}
where ${\bf X} \in \mathbb{R}^{I_1\times I_2}$ and elements of ${\bf X}$ are determined such that the rank of the matrix ${\bf X}$ is as small as possible. However, finding the rank of a matrix is non-convex therefore the optimization problem in \eqref{rank_minimization} becomes a non-convex problem. One common approach to solve this is to use the matrix nuclear norm $\left \| . \right\|_{\ast}$ to approximate the rank. 
%The advantage ofthe trace norm is that k:kis the tightest convex envelop for the rank of matrices. This leads to the following convex optimization problem for matrix completion
%However, finding the rank  of a tensor is NP \cite{(Hillar and Lim 2013; Kolda and Bader 2009), work in (Liu et al. 2013)},  A paper by \cite{jin et al(),Signoretto et al. 2014)} extends the matrix trace norm to the tensor case. 
%Matrix norm has widely been used to replace the matrix rank and has been successful in reconstruction and completion tasks \cite{xu2021fast,lu2019tensor, yuan2018tensor,xue2018low,jin2015annihilating,liu2012tensor}. The trace norm has the advantage of being the tightest convex envelope for the rank of matrices \cite{lu2019tensor}.
The nuclear norm has the advantage of being the tightest convex envelop for the rank of matrices \cite{bach2008consistency,cai2010singular,liu2012tensor}.
Matrix nuclear norm has widely been used  and has been successful in reconstruction and completion tasks \cite{yuan2018tensor}. Adopting nuclear  norm the problem in \eqref{rank_minimization} becomes
\begin{equation}\label{MinProSpa}
\begin{array}{l}
\mathop {\min }\limits_{{\bf X} }
\,\,\,\,\,    {{{\left\| {{\bf X}} \right\|}_\ast}} \\
%+ \left\| {{{\bf P}_{{\Omega}}}\left( {{\bf X} -  {{\bf Y} }}  \right)} \right\|_F^2\, 
s.t. \,\,\,\, {\bf X}_{\Omega}={\bf Y}_{\Omega},
\end{array}
\end{equation}
with $\left\| . \right\|_\ast$ representing the nuclear norm of a matrix which is the sum of the singular values of the matrix.

To use the matrix completion formulation \eqref{MinProSpa}, when we are dealing with data tensors, we can use the unfolding of the underlying data tensor and replaced it in \eqref{MinProSpar} ($ {\bf X}_{(1)}\leftarrow \underline{\bf X},\, {\bf Y}_{(1)}\leftarrow \underline{\bf Y} $ and ${\bf \Omega}_{(1)} \leftarrow \underline{\bf \Omega}$).
%From our optimization problem in \eqref{MinProb}, we transform the tensor into a matrix using the matrix unfolding: $\underline{\bf X} \rightarrow {\bf X}_{(1)}$ and $\underline{\bf Y} \rightarrow {\bf Y}_{(1)}$. 
Therefore we obtain the following optimization problem
\begin{equation}\label{MinProSpar}
\begin{array}{l}
\mathop {\min }\limits_{{\bf X}_{(1)}}
\,\,\,\,\,    {{
{
\lambda \, \left\| {{\bf X}_{(1)}} \right\|}_\ast}
}  + 
\left\| 
{\mathcal{P}_{{\bf \Omega}_{(1)}}}
\left( 
{\bf X}_{(1)} -  {{\bf Y}_{(1)} }   
  \right) \right\|_F^2,\,
s.t. \,\,\,\, {\bf X}_{{\bf \Omega}_{(1)}}={\bf Y}_{{\bf \Omega}_{(1)}}
\end{array}
\end{equation}
where $\lambda$ denotes the trade-off parameter. A constrained optimization problem can be formulated by introducing an auxiliary matrix ${\bf Z}_{(1)}$ with the same size as the matrix ${\bf X}_{(1)}$
\begin{equation}\label{MinProSpar_aux}
\begin{array}{l}
\mathop {\min }\limits_{ {\bf X}_{(1)},{\bf Z}_{(1)} }
\,\,\,\,\,    {\lambda \,{{\left\| {{\bf X}}_{(1)} \right\|}_\ast}} + \left\| {{\mathcal{P}_{{\bf \Omega}_{(1)}}}\left( {{\bf Z}_{(1)} -  {{\bf Y}_{(1)} }}  \right)} \right\|_F^2\,  \\
\,\,\,\,\,\,\,\,\,\,\,\quad s.t.\quad\quad{{\bf X}_{(1)}} = {{\bf Z}_{(1)}}.
\end{array}
\end{equation}
For the brevity of presentation, we use the notation ${\bf X}_{(1)} = {\bf X}$, ${\bf Z}_{(1)} = {\bf Z}$, ${\bf Y}_{(1)} = {\bf Y}$.
%In the rest of the paper, all matrix unfoldings are written in its matrix form. 
%That is: ${\bf X}_{(1)} = {\bf X}$, ${\bf Z}_{(1)} = {\bf Z}$, ${\bf Y}_{(1)} = {\bf Y}$.
We solve the minimization problem \eqref{MinProSpar_aux} via the Alternating Direction Method of Multipliers (ADMM) algorithm \cite{boyd2011distributed} which has been shown to have fast convergence and good performance. 
%The augmented Lagrangian function corresponding to the constrained optimization problem \eqref{MinProSpar_aux}, can be constructed as
\textcolor{black}{To do so, the augmented Lagrangian function for the constrained optimization problem \eqref{MinProSpar_aux} is first written as}
\begin{equation}\label{EQ_lagrangian}
\mathcal{L}\left( { {{\bf X}},{{\bf Z}},{{\bf T}} } \right) =  \lambda \left\| {{\bf X}} \right\|_\ast +
\left\| {{\mathcal{P}_{{\bf \Omega}_{(1)}}}\left( {{\bf Z} -  {{\bf Y} }}  \right)} \right\|_F^2  + 
\left\langle { {{\bf T}},{{\bf X}} - { {{\bf Z}}} } \right\rangle +\frac{\mu}{2}\left\| {  {{\bf X}} - {{\bf Z}}  } \right\|_F^2,
\end{equation}
\textcolor{black}{where ${{\bf T}}$ is a matrix representing the Lagrangian multipliers and $\mu$ is a penalty parameter.}
In our simulation results the ADMM method worked properly for $\mu$ between 0.1 to 1.1.
%\textcolor{blue}{All hyper-parameters were optimized to provide the best performance}.
%Based on Lagrangian function \eqref{EQ_lagrangian}, the ADMM \cite{boyd2011distributed} update rules  for solving \eqref{MinProSpar_aux} are straightforwardly converted to simpler optimization problems as follows:
According to the ADMM method, we update the matrices ${\bf X},\,{\bf Y},\,{\bf Z}$, iteratively by fixing two of them and updated the other as presented below.

\textbf{Update of ${\bf X}$:} By minimizing the augmented Lagrangian function \eqref{EQ_lagrangian} w.r.t. ${\bf X},$ we have
\begin{eqnarray}\label{lagragian_subprob1}
{{\bf X}_{k+1}} = \mathop {\min }\limits_{{\bf X}} \,\, \mathcal{L}
\left(  {{\bf X}}, {{\bf Z}}_k , {{\bf T}}_k, {\mu_k}   \right),
\end{eqnarray}
%The first optimization sub-problem \ref{lagragian_subprob1} for $\underline{\bf X}$ can be solved via
which can be further simplified to 
%\ref{lagragian_subprob1} for $\underline{\bf X}$ can be solved via
\begin{equation}\label{subprob1}
  {{\bf X}_{k+1}} = \mathop {\min }\limits_{{\bf X}} \,\, \left( \lambda \left\| {{\bf X}} \right\|_\ast + \frac{\mu_k}{2} \left\| {{\bf X}} - {{\bf Z}}_k + \frac{{\bf T}_k}{\mu_k} \right\|_F^2  \right).  
\end{equation}
According to the paper \cite{cai2010singular}, the above problem has the closed form solution given by:
\begin{equation}\label{subprob1_solved}
    {{\bf X}_{k+1}} = \bf D_\beta\left( {{\bf Z}}_k - \frac{{\bf T}_k}{\mu_k} \right),
\end{equation}
where ${\beta}>0$ is constant and $\bf D_{\beta}(.)$ is the matrix singular value thresholding operation defined as follows
\begin{equation}
    {\bf D}_\beta({\bf X})  = {{\bf U} \ast {\bf D}_\beta({\bf S}) \ast {\bf V}}^{T},
\end{equation}
where $\underline{\bf D}_\beta({\bf S}^{(i)})=$  $ {\rm diag}(\max\{\sigma_t -\beta,0  \}_{1\leq t\leq R}), \, i=1,\ldots, I_3$, $\beta>0$ is a constant and $R$ is the rank of ${\bf S}$. Here, the ${\bf X}={{\bf U} {\bf S} {\bf V}}^{T}$ SVD of the matrix ${\bf X}$. 
%where  $\bf D_{\frac{\lambda}{\mu_k}}(.) $ is the matrix singular value thresholding operation. That is, ${{\bf U} {\bf S} {\bf V}}^{T}$ is the singular value decomposition of a matrix ${\bf A}$. The sub-problem \ref{lagragian_subprob2} for ${\bf Z}$ can be solved through:

\textbf{Update of ${\bf Z}$:} By minimizing the augmented Lagrangian function w.r.t. ${\bf Z}$, we have
\begin{eqnarray}\label{lagragian_subprob2}
{{\bf Z}_{k+1}} = \mathop {\min }\limits_{{\bf Z}} \,\, \mathcal{L}
\left(  {{\bf X}}_{k+1}, {{\bf Z}} , {{\bf T}}_k, {\mu_k}   \right) ,
\end{eqnarray}
and can be simplified as 
%The sub-problem \ref{lagragian_subprob2} for ${\bf Z}$ can be solved through:
\begin{equation}\label{subprob2}
{{\bf Z}_{k+1}} = \mathop {\min }\limits_{{\bf Z}_k^{(n)}} \,\, \left\| {{\mathcal{P}_{{\bf \Omega}}}\left( {{\bf Z} -  {{\bf Y} }}  \right)} \right\|_F^2 + \frac{\mu_k}{2} \left\| {{\bf X}}_{k+1} - {{\bf Z}} + \frac{{\bf T}_k}{\mu_k} \right\|_F^2
\end{equation}
The closed form solution for problem \ref{subprob2} is also solved through
\begin{equation}\label{subprob2_solved}
    {{\bf Z}_{k+1}} = {\mathcal{P}_{{\bf \Omega}}^{\bot}} \left( {{\bf X}}_{k+1} + \frac{{\bf T}_k}{\mu_k}  \right) + {{\bf Y}}
\end{equation}

\textbf{Update of ${\bf T}$:} The solution for the Lagrangian multiplier matrix ${\bf T}$ is similarly converted to simpler optimization problem as follows
\begin{eqnarray}\label{lagragian_subprob3}
{{\bf T}_{k + 1}} = \mathop {\min }\limits_{{\bf T}} \,\, \mathcal{L}
\left(  {{\bf X}}_{k+1}, {{\bf Z}}_{k+1} , {{\bf T}}, {\mu_k}   \right).
\end{eqnarray}
which has the close solution
\begin{equation}\label{subprob3_solved}
{{\bf T}_{k + 1}} = {{\bf T}_{k}} + \mu_k \left( {{\bf X}_{k+1} -  {{\bf Z}_{k+1}} } \right).
\end{equation}
Besides, we update the parameter ${\mu}$ in the following way
\begin{eqnarray}\label{lagragian_subprob4}
{{\bf \mu}_ {k + 1}} = \mathop {\min } \,\, 
\left( \alpha{\mu_k}, {\mu_{max}}  \right).
\end{eqnarray}
where $\alpha > 1$ is a predetermined constant used to iteratively increase the penalty and $\mu_{max}$ represents the upper bound for the penalty. This procedure is summarized in Algorithm \ref{ALG:SMNN}. As we will discuss in Subsection \ref{Sec:smooth}, to improve the quality of the image reconstruction process, we smooth the auxiliary matrix ${\bf Z}$ in Line of Algorithm \ref{ALG:SMNN} after its computation. This smoothing approach totally improves the results as will be shown in the simulation part. 
%\textcolor{blue}{It is worth to note that the ADMM algorithm is a workhorse approach for solving a variety of optimization problems such as sparse inverse covariance selection, generalized and group lasso problems etc., for a comprehensive study of this technique and related problems, see \cite{boyd2011distributed}.
%Moreover, this strategy has been utilized for solving many tensor completion problems, please see \cite{huang2020robust,yang2018tensor,yuan2020rank,zhao2019low,ding2020tensor}, and for a comprehensive list of references see the review papers \cite{song2019tensor,long2019low} and the references therein.}

\RestyleAlgo{ruled}
\begin{algorithm}
\LinesNumbered
\SetKwInOut{Input}{Input}
\SetKwInOut{Output}{Output}\Input{An observed data tensor $\underline{\bf Y} \in \Real^{I_1 \times I_2\times \cdots \times I_N}$, the observation index tensor $\underline{\bf \Omega}$, and regularization parameter $\lambda > 0$, ${\bf X}_0 ={\bf T}_0 $, ${\bf Z}_0 ={\bf X}_0 $, ${ {\mathcal{P}_{{\bf \Omega}} } \left({\bf X} \right) = {\mathcal{P}_{{\bf \Omega}}} \left({{\bf Y}} \right) }$. }
\Output{Completed data tensor ${\underline{\bf X}}$}
\caption{Algorithm for Smooth Matrix Nuclear Norm (SMNN)}\label{ALG:SMNN}
      \textbf{Perform} Super-pixel extraction to generate compressed data  \\
      \textbf{Perform} mode-n unfolding of tensor ${\underline{\bf X}}$\\
\While{A stopping criterion is not satisfied}
{
       	Update ${{\bf X}}_{k+1} $  using solution from equation \ref{subprob1_solved} \\
       	Update ${{\bf Z}}_{k+1} $ using solution from \ref{subprob2_solved} \\
       	Perform Smoothing operation \\
       	Update ${{\bf T}}_{k+1} $  using equation in  \ref{subprob3_solved} \\
       	Update ${{\mu}}_{k+1} $  using solution from equation \ref{lagragian_subprob4} \\
    %   Check convergence conditions  \\
    %   $\left\| { {{\bf X}_{k+1} -  {{\bf X}_{k} }}  } \right\|_\ast \leq tol$ \\
 }  
\textbf{Reshape} ${\bf X}$ into tensor ${\underline{\bf X}}$ \\
\textbf{Compute} ${\underline{\bf X}} =  \mathcal{P}_{\underline{\bf \Omega}} \left( {\underline{\bf X}} \right) + \mathcal{P}_{{\underline{\bf \Omega} }^{\perp}} \left( {\underline{\bf X}} \right)$  \\
\textbf{Return} ${\underline{\bf X}}$ 
\end{algorithm}

\subsection{Tensor completion using tensor nuclear norm (TNN) regularization}\label{SEC:TA}  
In this section, we discuss the tensor formulation of the completion problem based on the tensor Singular Decomposition (t-SVD) model. The motivation for this new formulation is comparing the performance of the matrix and the tensor variants in reconstructing the images. We have seen better results of the tensor case than the matrix case as will be discussed in the simulation section. Kilmer et al. \cite{kilmer2013third} proposed the t-SVD as a new tensor decomposition, for detailed description of this tensor model, see the Appendix. Inspired by the results achieved by the nuclear norm minimization of matrices for recovering data matrices with missing values, Zhang \textit{et al.}\cite{zhang2016exact} proposed the tubal nuclear norm minimization approach based on t-SVD, defined as the sum of nuclear norms of all frontal slices in the Fourier domain and proved to be convex envelope to the tensor tubal rank (See the Appendix for the details). 
%Abstracting the benefits of these methods, we exploit and evaluate the prowess of the superpixel clustering method to tensor nuclear norm case for completion. 
More precisely, the model of rank minimization based tensor completion is formulated as follows
\begin{equation}\label{rank_minimization_tensor}
\begin{array}{l}
\mathop {\min }\limits_{ \underline{\bf X} }
\,\,\,\,\,    {{ {\rm tubal\,rank} \, ({\underline{\bf X}} }} )+ \left\| {{\mathcal{P}_{\underline{\Omega}}}\left( {\underline{\bf X} -  {\underline{\bf Y} }}  \right)} \right\|_F^2.\, 
\end{array}
\end{equation}
%However, finding the rank of a tensor is NP \cite{hillar2013most, kolda2009tensor,liu2012tensor}. 
Similar to the matrix case, minimizing the tubal tensor rank is NP hard because it includes the matrix case as a special case. The matrix trace norm was generalized to the tensor case based on the t-product in \cite{lu2019low,lu2019tensor,xue2018low,zhang2016exact}. We use the one introduced in \cite{xue2018low,lu2019tensor} which has been shown to provide superior results compared to the others and to be faster because of using only the information of the first slice in the Fourier domain. So, we consider the following minimization problem
%adopt the general completion model implemented in \cite{liu2012tensor} for the matrix case to higher order tensors shown in \cite{xu2021fast} by solving 
\begin{equation}\label{MinProSpar_tensor}
\begin{array}{l}
\mathop {\min }\limits_{ \underline{\bf X} }
\,\,\,\,\,    {\lambda \,{{\left\| {\underline{\bf X}} \right\|}_\ast}} + \frac{1}{2} \, \left\| {{\mathcal{P}_{\underline{\Omega}}}\left( {\underline{\bf X} -  {\underline{\bf Y} }}  \right)} \right\|_F^2,
\end{array}
\end{equation}
where $\left\| . \right\|_\ast$ is the tubal nuclear norm 
%sum of the singular values of the tensor referred to as the trace norm or nuclear norm 
and $\lambda$ denotes the trade-off parameter. Note that the truncated tubal nuclear norm \cite{xue2018low} can also be used in the formulation \eqref{MinProSpar_tensor}.
Similar tensor completion formulation is used in \cite{xu2021fast} but here we have used  unitary transform matrices instead of discrete
Fourier transform matrix that is used in the traditional tensor SVD and has shown to provide better results \cite{song2020robust}. We also proposed to improve the image recovery by smoothing the results at each iteration (Line 5 in Algorithm \ref{ALG:STNN}). To use the ADMM algorithm, we need to  introduce an auxiliary tensor $\underline{\bf Z}$ with same size as the tensor $\underline{\bf X}$: 
\begin{equation}\label{MinProSpar_aux_tensor}
\begin{array}{l}
\mathop {\min }\limits_{ \underline{\bf X},\underline{\bf Z} }
\,\,\,\,\,    {\lambda \,{{\left\| {\underline{\bf X}} \right\|}_\ast}} + \frac{1}{2} \,\left\| {{\mathcal{P}_{\underline{\bf \Omega}}}\left( {\underline{\bf Z} -  {\underline{\bf Y} }}  \right)} \right\|_F^2\,  \\
\,\,\,\,\,\,\,\,\,\,\,\quad s.t.\quad\quad{\underline{\bf X}} = {\underline{\bf Z}},
\end{array}
\end{equation}
Here again, the augmented Lagrangian function corresponding to the constrained optimization problem \eqref{MinProSpar_aux_tensor}, is written as
\begin{equation}\label{EQ_lagrangian_tensor}
\mathcal{L}\left( { {\underline{\bf X}},{\underline{\bf Z}},{\underline{\bf T}} } \right) =  \lambda \left\| {\underline{\bf X}} \right\|_\ast + \frac{1}{2} \,
\left\| {{\mathcal{P}_{\underline{\bf \Omega}}}\left( {\underline{\bf Z} -  {\underline{\bf Y} }}  \right)} \right\|_F^2  + 
\left\langle { {\underline{\bf T}},{\underline{\bf X}} - { \underline{{\bf Z}}} } \right\rangle  + \frac{\mu}{2}\left\| {  {\underline{\bf X}} - {\underline{\bf Z}}  } \right\|_F^2,
\end{equation}
where $\underline{{\bf T}}$ is a tensor representing the  Lagrangian multipliers and $\mu$ is a penalty parameter.
%While the method works well for $\mu$  between 0.1 to 1.1. We set $\lambda=0.5$ in our simulations.  
%\textcolor{blue}{All hyper-parameters were optimized to provide the best performance}.
Similarly as done for the matrix case, the Lagrangian function \eqref{EQ_lagrangian_tensor}, is minimized with respect to the tensors ${\underline{\bf X}},\,{\underline{\bf Y}},{\underline{\bf T}},$ by fixing two of them and updating the other.
%and the ADMM \cite{boyd2011distributed} solution is converted to simpler sub-problems
Let us start with the tensor ${\underline{\bf X}}$ and by minimizing the augmented Lagrangian function \eqref{EQ_lagrangian_tensor} with respect to the tensor ${\underline{\bf X}}$, we have
%sub-problem for $\underline{\bf X}$ is updated as
\begin{eqnarray}\label{Tensor_lagragian_subprob1}
{\underline{\bf X}_{k+1}} = \mathop {\min }\limits_{\underline{\bf X}} \,\, \mathcal{L}
\left(  {\underline{\bf X}}, {\underline{\bf Z}}_k , {\underline{\bf T}}_k, {\mu_k}   \right), 
\end{eqnarray}
which can be simplified as
\begin{equation}\label{Tensor_subprob1}
  {\underline{\bf X}_{k+1}} = \mathop {\min }\limits_{\underline{\bf X}} \,\, \left( \lambda \left\| {\underline{\bf X}} \right\|_\ast + \frac{\mu_k}{2} \left\| {\underline{\bf X}} - {\underline{\bf Z}}_k + \frac{\underline{\bf T}_k}{\mu_k} \right\|_F^2  \right).  
\end{equation}
The minimization problem \eqref{Tensor_lagragian_subprob1}, has the close form solution 
\begin{equation}\label{Tensor_subprob1_solved}
    {\underline{\bf X}_{k+1}} = \underline{\bf D}_\beta\left( {\underline{\bf Z}}_k - \frac{\underline{\bf T}_k}{\mu_k} \right),
\end{equation}
where  $\underline{\bf D}_\beta(.) $ is the tensor singular value thresholding operation and defined similar to the matrix case (See the Appendix). 
%That is, ${\underline{\bf U} \underline{\bf S} \underline{\bf V}}^{T}$ is the singular value decomposition of a tensor $\underline{\bf A}$. 
To update $\underline{\bf Z}$, consider
%The augmented Lagragian function w.r.t $\underline{\bf Z}$ is updated by solving the minimization problem
\begin{eqnarray}\label{Tensor_lagragian_subprob2}
{{\bf Z}_{k+1}} = \mathop {\min }\limits_{{\bf Z}} \,\, \mathcal{L}
\left(  {\underline{\bf X}}_{k+1}, {\underline{\bf Z}} , {\underline{\bf T}}_k, {\mu_k}   \right),
\end{eqnarray}
which be simplified as 
\eqref{Tensor_lagragian_subprob2} for $\underline{\bf Z}$ can be solved through:
\begin{equation}\label{Tensor_subprob2}
{{\bf Z}_{k+1}} = \mathop {\min }\limits_{{\bf Z}_k} \,\,\frac{1}{2} \, \left\| {{\mathcal{P}_{\underline{\bf\Omega}}}\left( {\underline{\bf Z} -  {\underline{\bf Y} }}  \right)} \right\|_F^2 + \frac{\mu_k}{2} \left\| {\underline{\bf X}}_{k+1} - {\underline{\bf Z}} + \frac{\underline{\bf T}_k}{\mu_k} \right\|_F^2.
\end{equation}
and has the closed form solution defined as 
\begin{equation}\label{Tensor_subprob2_solved}
    {{\bf Z}_{k+1}} = {\mathcal{P}_{\underline{\bf\Omega}}^{\bot}} \left( {\underline{\bf X}}_{k+1} + \frac{\underline{\bf T}_k}{\mu_k}  \right) + {\underline{\bf Y}}.
\end{equation}
The solution for $\underline{\bf T}$ is also converted to a simpler optimization as
\begin{eqnarray}\label{Tensor_lagragian_subprob3}
{\underline{\bf T}_{k + 1}} = \mathop {\min }\limits_{{\bf T}} \,\, \mathcal{L}
\left(  {\underline{\bf X}}_{k+1}, {\underline{\bf Z}}_{k+1} , {\underline{\bf T}}, {\mu_k}   \right).
\end{eqnarray}
which can be solved through:
\begin{equation}\label{Tensor_subprob3_solved}
{\underline{\bf T}_{k + 1} } = {\underline{\bf T}_{k}} + \mu_k \left( {\underline{\bf X}_{k+1} -  {\underline{\bf Z}_{k+1}} } \right),
\end{equation}
Similar to the matrix case, we update the penalty parameter ${\bf \mu}$ via
\begin{eqnarray}\label{Tensor_lagragian_subprob4}
{{\bf \mu}_{k + 1}} = \mathop {\min } \,\, 
\left( \mathcal{J}{\mu_k}, {\mu_{max}}  \right).
\end{eqnarray}

\RestyleAlgo{ruled}
\begin{algorithm}
\LinesNumbered
\SetKwInOut{Input}{Input}
\SetKwInOut{Output}{Output}\Input{An observed data tensor $\underline{\bf Y} \in \Real^{I_1 \times I_2\times \cdots \times I_N}$, the observation index tensor $\underline{\bf \Omega}$, and regularization parameter $\lambda > 0$, $\underline{\bf X}_0 =\underline{\bf T}_0 $, $\underline{\bf Z}_0 =\underline{\bf X}_0 $, ${ {\mathcal{P}_{\underline{\bf \Omega}} } \left( \underline{\bf X} \right) = {\mathcal{P}_{\underline{\bf \Omega}}} \left({\underline{\bf Y}} \right) }$. }
\Output{Completed data tensor ${\underline{\bf X}}$}
\caption{Algorithm for Smooth Tensor Nuclear Norm (STNN)}\label{ALG:STNN}
      \textbf{Perform} Super-pixel extraction to generate compressed data  \\

\While{A stopping criterion is not satisfied}
{
       	Update ${\underline{\bf X}}_{k+1} $  with equation \ref{Tensor_subprob1_solved} \\
       	Update ${\underline{\bf Z}}_{k+1} $  with equation \ref{Tensor_subprob2_solved} \\
       	Perform Smoothing operation \\
       	Update ${\underline{\bf T}}_{k+1} $  with equation \ref{Tensor_subprob3_solved} \\
       	Update ${{\mu}}_{k+1} $  using equation \ref{Tensor_lagragian_subprob4} \\
      Check convergence conditions  \\
      $\left\| { {\underline{\bf X}_{k+1} -  {\underline{\bf X}_{k} }}  } \right\|_\ast \leq tol$ \\
\textbf{Compute} ${\underline{\bf X}} = {\mathcal{P}_{\underline{\bf \Omega}} }\left( \widehat{\underline{\bf X}}_H \right) + \mathcal{P}_{{\underline{\bf \Omega} }^{\perp}} \left( {\widehat{\underline{\bf X}}} \right)$  \\
%\textbf{Return} ${\underline{\bf X}}$ 
}  
\end{algorithm}

\begin{figure*}[ht!]
    \centering
    \includegraphics[width=0.75\linewidth]{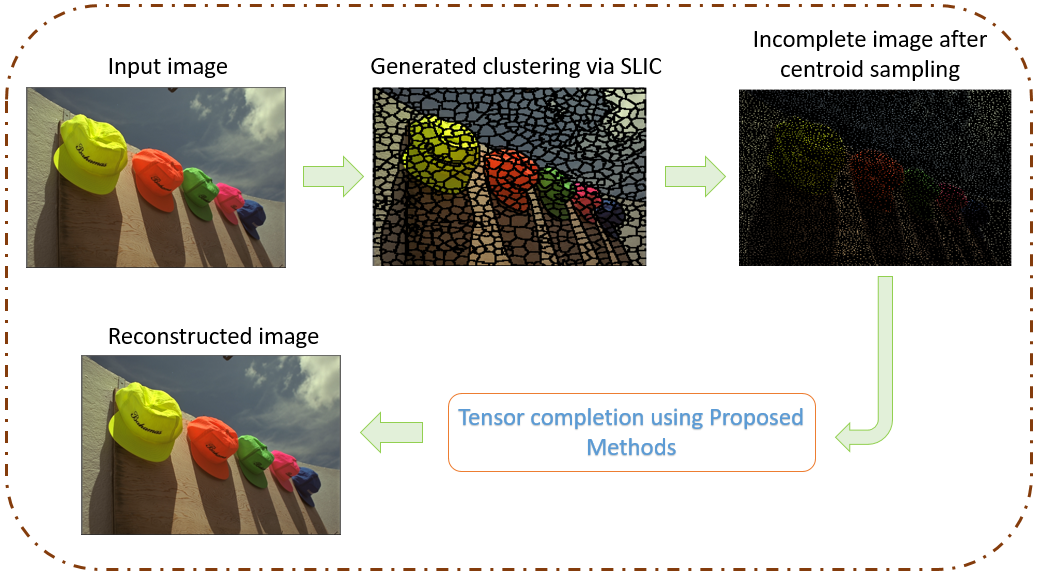} 
    \caption{Pipeline of proposed method}
    \label{fig:flowchart_pipeline}
\end{figure*}

\subsection{Smoothing techniques}\label{Sec:smooth}
\textcolor{black}{Low rank modeling of data has achieved tremendous success in tensor completion, however, only a low rank prior is inadequate for a successful recovery of the underlying tensor \cite{yokota2016smooth,zheng2019low,he2019remote}. The case is much difficult when the number of missing pixels are high. Thankfully, many real world images and data exhibit some smoothness prior along both the spatial and the third modes especially in the case of RGB images, videos, and hyperspectral images \cite{zheng2019low, he2019total}. As such, it becomes a very useful property in modelling these types of data. The assumption of smoothness in data means that the differences between neighboring values are small in certain domains. For example, non-negative natural images are smooth in the spatial domain. Therefore, the smoothing constraint is a common assumption used to improve results when dealing with some datasets such as images and videos. More precisely, the smoothness constraint is imposed on the underlying factors along with the low-rank assumption of the tensor decomposition used in reconstruction model. Indeed, matrix/tensor factorization methods with smoothness constraints have a wide range of applications that require robustness in the presence of noisy signals, including  image in-painting, denoising, brain signal analysis and hyperspectral imaging. As a result, many tensor completion algorithms have been proposed imposing smoothness on the recovered data \cite{yokota2017simultaneous,zheng2019low, sinha2022nonnegative,yokota2016smooth,ahmadi2022cross,ahmadi2022cross}.
A very popular regularizer used to impose piece-wise smoothness is the Total Variation (TV) \cite{rudin1992nonlinear}. The TV is determined by the $l_1$-norm  of the difference between neighboring elements. Many methods in matrix and tensor completion have used the TV approach \cite{yokota2016smooth, yokota2016tensor, ding2019low},\cite{zheng2019low, zheng2020tensor, he2019remote,he2019total,ji2016tensor}. The works by \cite{yokota2016smooth} and \cite{li2017low} incorporates smoothness into a PARAFAC model for partially observed tensors. Both models created two variants for the completion task. The former proposed the models based on total variation and quadratic variation regularizes. The approach by \cite{li2017low} uses the total variation (TV) regularizer to also formulate the tensor completion model, taking advantage of a piecewise prior and local smoothness constraint. The approaches adopts Canonical Polyadic(CP) and  Tucker decomposition for a simultaneous decomposition and completion task. Furthermore, \cite{zheng2019low} adopts the tensor smoothness constraints using smooth matrix factorizations. The methods in \cite{cai2008framelet, ji2016tensor,jiang2018matrix, zheng2019low} exploit the spatial piecewise smoothness prior of the underlying tensor by increasing the piecewise smoothness of each row in the data.}
Other smoothness approaches such as methods used in \cite{ahmadi2022cross,ko2020fast},\cite{xue2021multilayer, xue2019enhanced,li2019low,jiang2018matrix} have demonstrated significant improvement in results using various low rank and higher order tensor networks.
In addition, different techniques and methodologies can be used for smoothing out the elements of data elements such as low-pass filtering, moving averaging, locally estimated scatterplot smoothing (LOESS) and gaussian filtering. We perform a simple Gaussian smoothing filtering with standard deviation which returns the filtered image. The ``imgaussfilt'' command in Matlab can be used for the mentioned smoothing technique.
%zhao2020deep,el2022hyperspectral
%In fact we experimentally found that the smooth constraint provides better results of other incomplete image

\section{Experiments}\label{Experiments:Sec}
\begin{figure*}[ht!]
    \centering
    \includegraphics[width=1.0\linewidth]{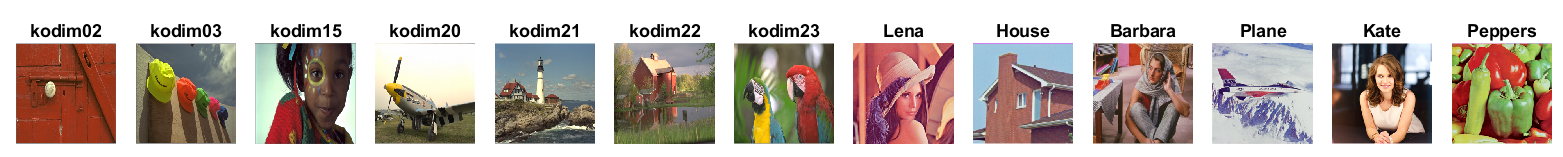} 
    \caption{Colored images used in numerical experiments.}
    \label{fig:benchmarkimages}
\end{figure*}

In this section, we present an evaluation of the proposed algorithm using real colored images. Experiments or simulations were performed on a laptop computer with 16GB memory and  a 2.60 GHz Intel(R) Core(TM) i7-5500U processor. The Peak Signal-to-Noise Ratio (PSNR) and  Structural Similarity Index measure (SSIM) metrics were used to evaluate and compare the performance of different algorithms. We mainly consider three experiments. In the first simulation, we extensively compare the proposed superpixel sampling approach with the random sampling and show that in all cases the superpixel method provide better results. In the second simulation, we will use the superpixel sampling to remove the pixels as was shown in the first simulation to be better than the random one and compared the proposed completion algorithms with the baseline completion algorithms. Here, the effect of smooth filtering in the reconstruction performance is also illustrated. In the third simulation, we consider the more challenging cases where the pixels are removed in structural way and again compared the proposed completion algorithms with the baseline algorithms. We have used the RGB images ``Kate'', ``House'', ``Lena'',``Plane'', and ``Peppers'' and also some images from the Kodak dataset \cite{kodakimages} as our benchmark images which are
 shown in Figure \ref{fig:benchmarkimages}. ``Lena'', ``House'', ``Peppers'' and ``Kate'' images are of size $256 \times 256 \times 3 $ whiles the Kodak images are of size $512 \times 768 \times 3 $. 
 %The value for $k$ in the super-pixel clustering calculated based on the sampling rate.

\begin{exa}\label{exa_1}
{\bf  (Comparison of superpixel sampling methods and random sampling.)}
This experiment is devoted to illustrating the superiority of the proposed superpixel sampling approach compared to the random sampling method which was mostly used in literature. In Section \ref{Sec:Tencom} (Figures \ref{fig:sampling_methods} and \ref{fig:centroid_vs uniform}), we showed the better performance of the superpixel sampling compared to random sampling for kodak images (kodim03, kodim23) and here we make new experiments for the kodim15, kodim22 and kodim02 images. Here, we make more experiments using images with $60\%$, $30\%$ and $20\%$ of pixel sampling. The pixels are sampled according to different superpixel sampling approaches described in Section \ref{Sec:Tencom} including Centroid, Boundary, Multi-stage sampling and also random sampling. Note that for the completion part (stage 2), we have used Algorithm \ref{ALG:STNN} (STNN) with smoothing filtering. The results are reported in Figure \ref{fig:Compare_sampling_methods}. As can be seen, in all scenarios, the centriod superpixel sampling provided the best results. These extensive simulations on a variety of images and using different categories clearly convinced us that the centroid superpixel approach can select better pixels than other superpixel approaches and also the random sampling method. 

\begin{figure*}[ht!]
\centering
\subfigure[Reconstructed images with 60\% of pixels sampled from the kodim15 image using different sampling techniques.]{
\includegraphics[width=0.8\linewidth]{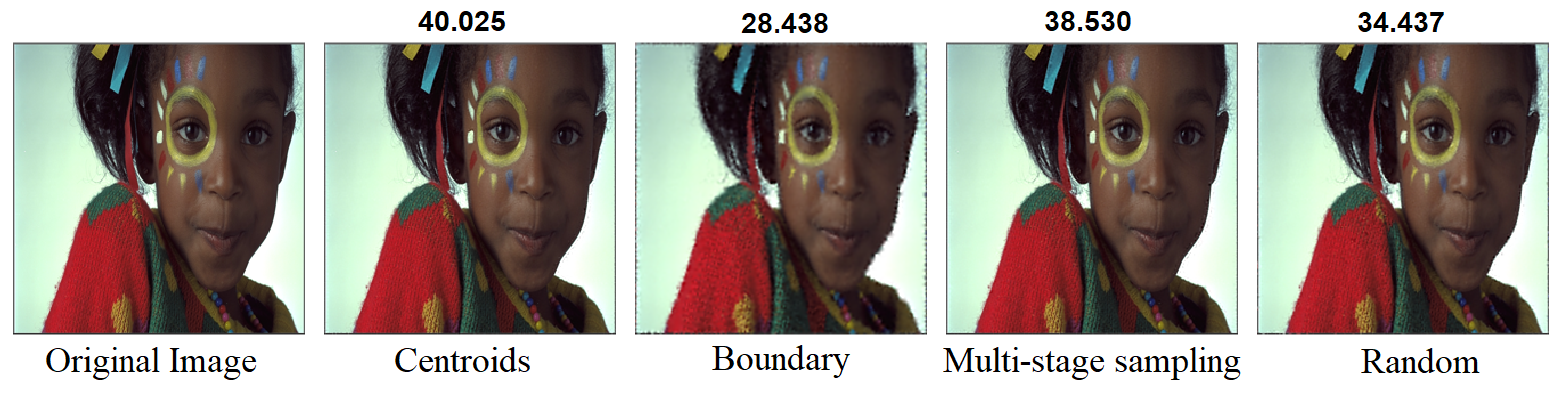}
}
\subfigure[Reconstructed images with 30\% of pixels sampled from the kodim23 image using different sampling techniques.]{
\includegraphics[width=0.8\linewidth]{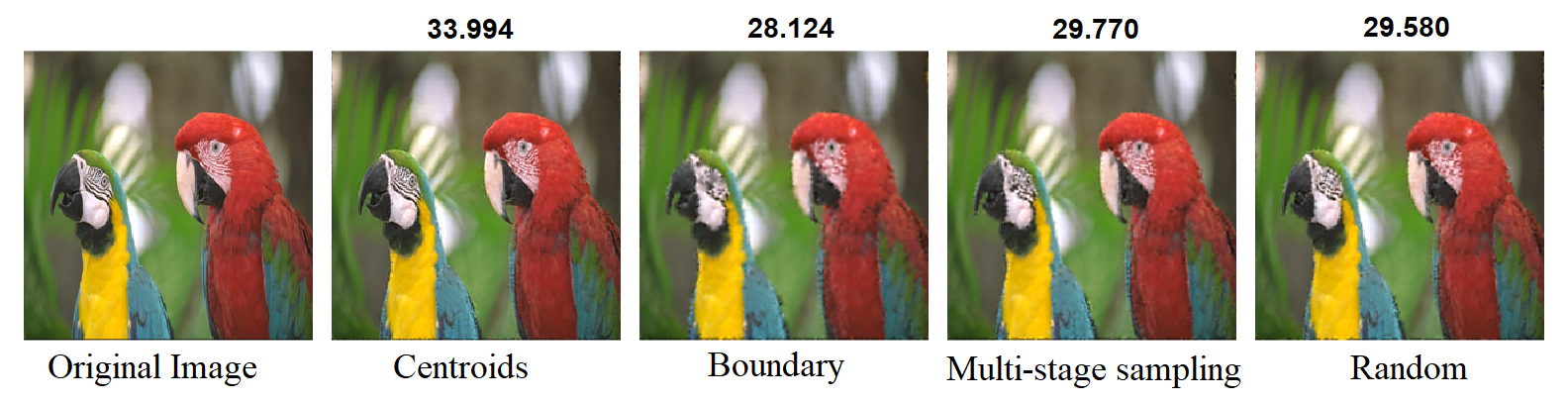}
}
\subfigure[Reconstructed images with 30\% of pixels sampled from the kodim22 image using different sampling techniques.]{
\includegraphics[width=0.8\linewidth]{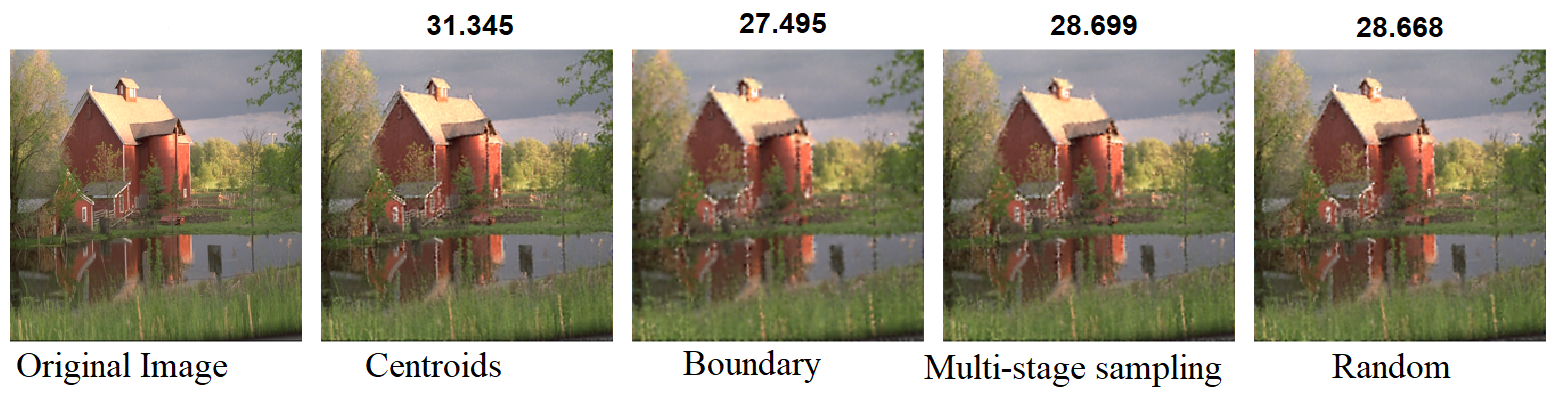}
}
\subfigure[Reconstructed images with 20\% of pixels sampled form the kodim02 image  using different sampling techniques.]{
\includegraphics[width=0.8\linewidth]{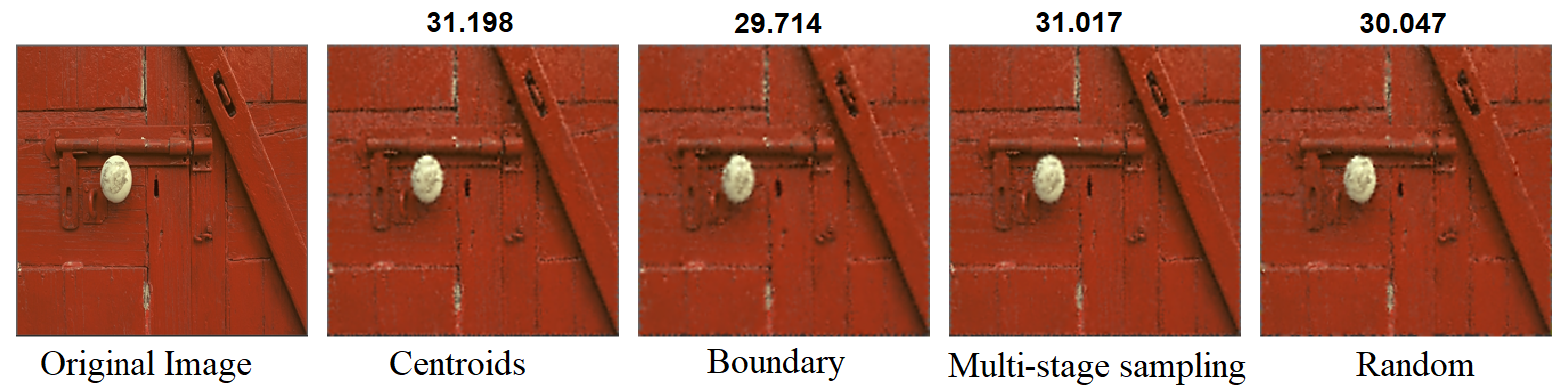}
}
\caption{PSNR comparison of Kodak images using different superpixels sampling methods for Example \ref{exa_1}. The experiment show that centroid pixel selection provides better results.}
\label{fig:Compare_sampling_methods}
\end{figure*}
\end{exa}

\begin{figure*}[ht!]
    \centering
    \includegraphics[width=1.0\linewidth]{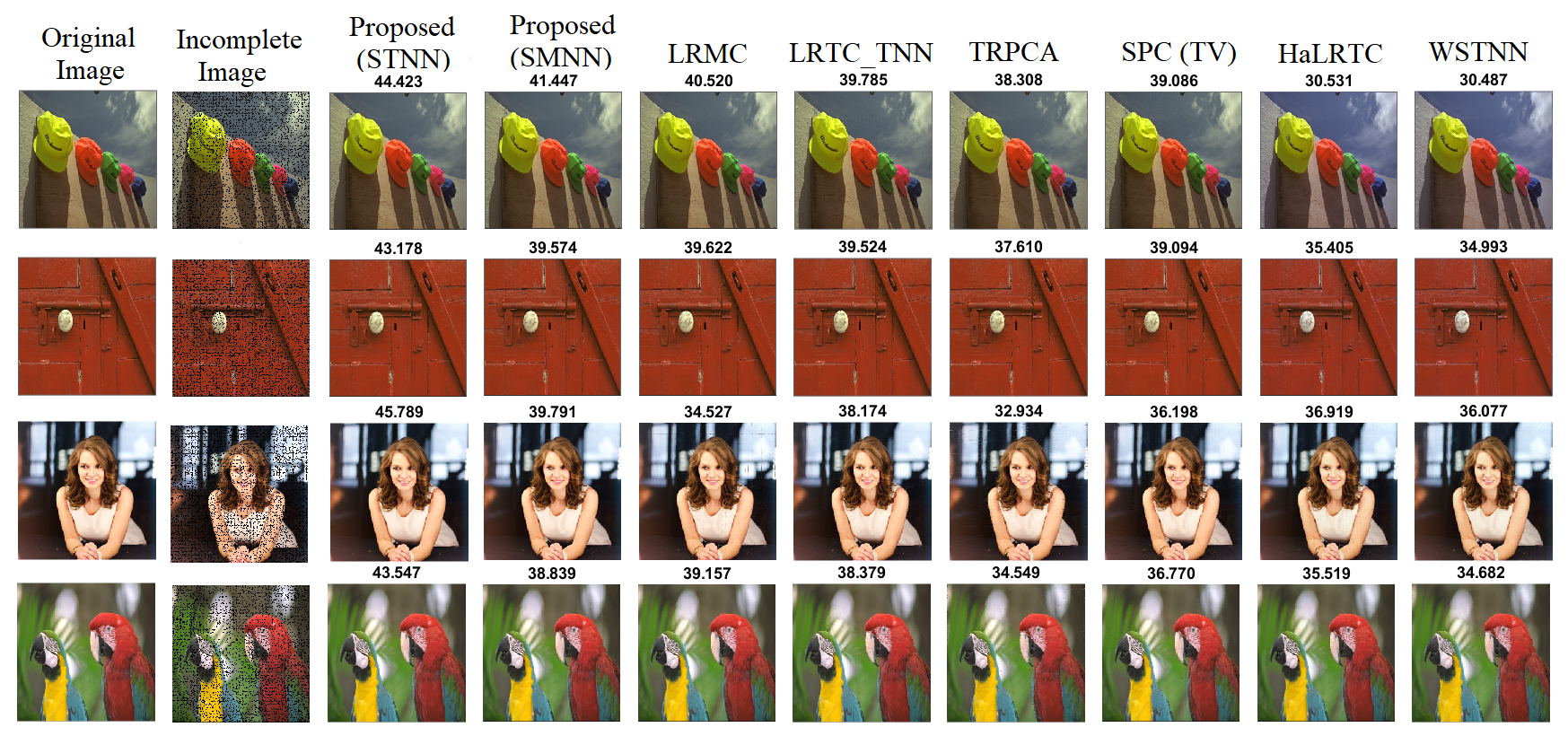} 
    \caption{PSNR comparison of different low rank completion methods with sampling 70\% of pixels for Example \ref{exa_2}.}
    \label{fig:30centroids}
\end{figure*}

\begin{figure*}[ht!]
    \centering
    \includegraphics[width=1.0\linewidth]{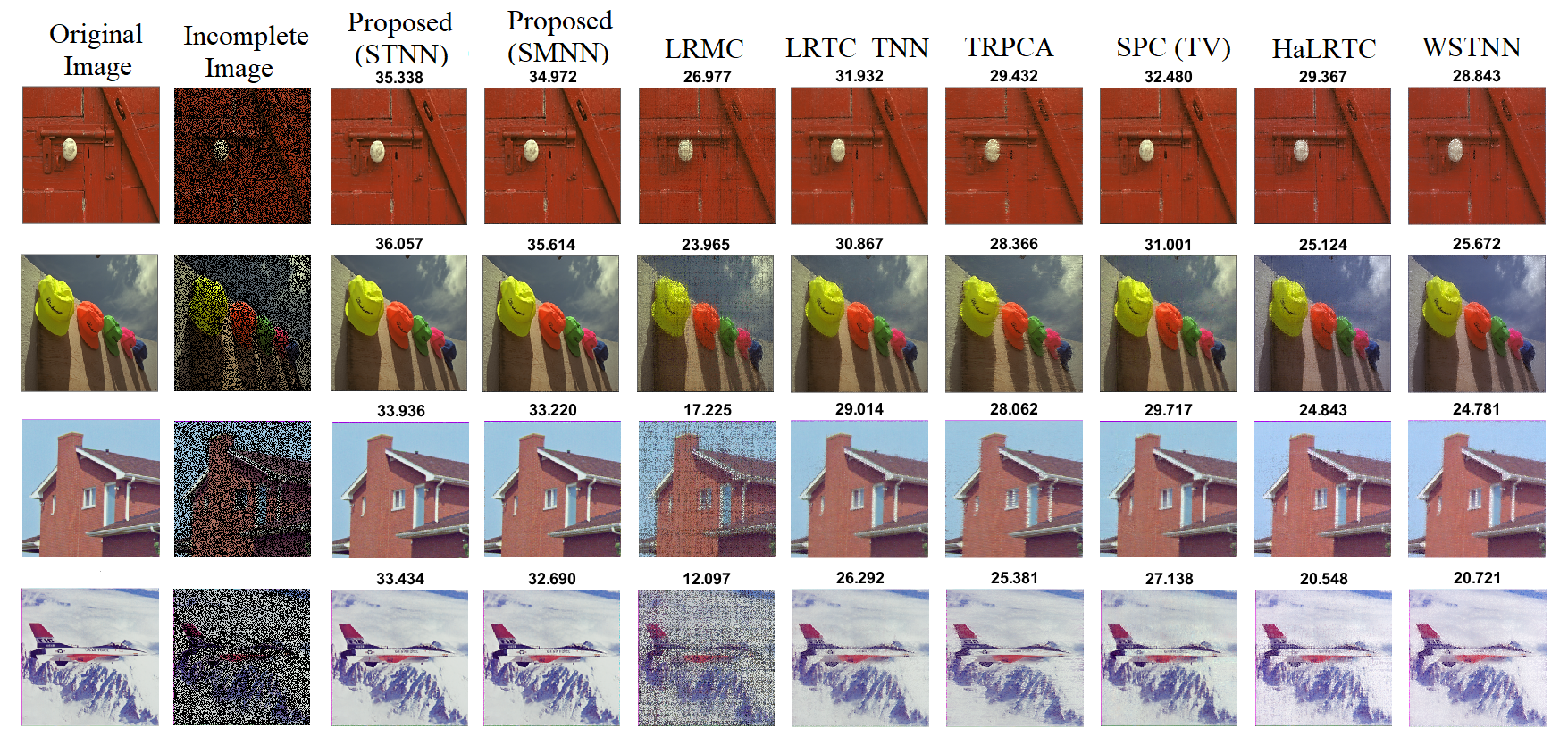} 
    \caption{PSNR comparison of different low rank completion methods with sampling 30\% of pixels for Example \ref{exa_2}.}
    \label{fig:70centroids}
\end{figure*}

\begin{figure*}[ht!]
    \centering
    \includegraphics[width=1.0\linewidth]{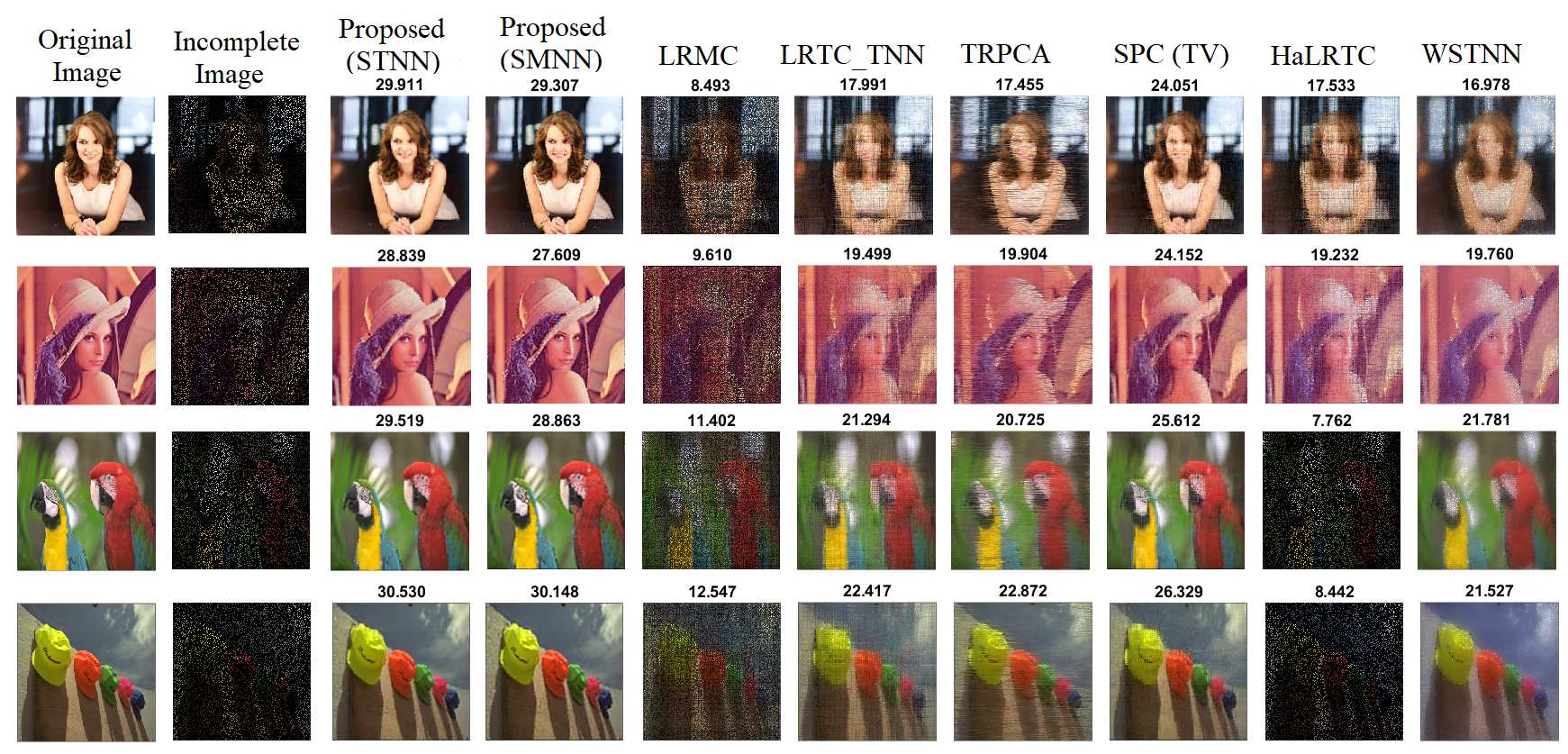} 
    \caption{PSNR comparison of different low rank completion methods with sampling 20\% of pixels for Example \ref{exa_2}.}
    \label{fig:80centroids}
\end{figure*}

\begin{figure*}[ht!]
    \centering
    \includegraphics[width=1.0\linewidth]{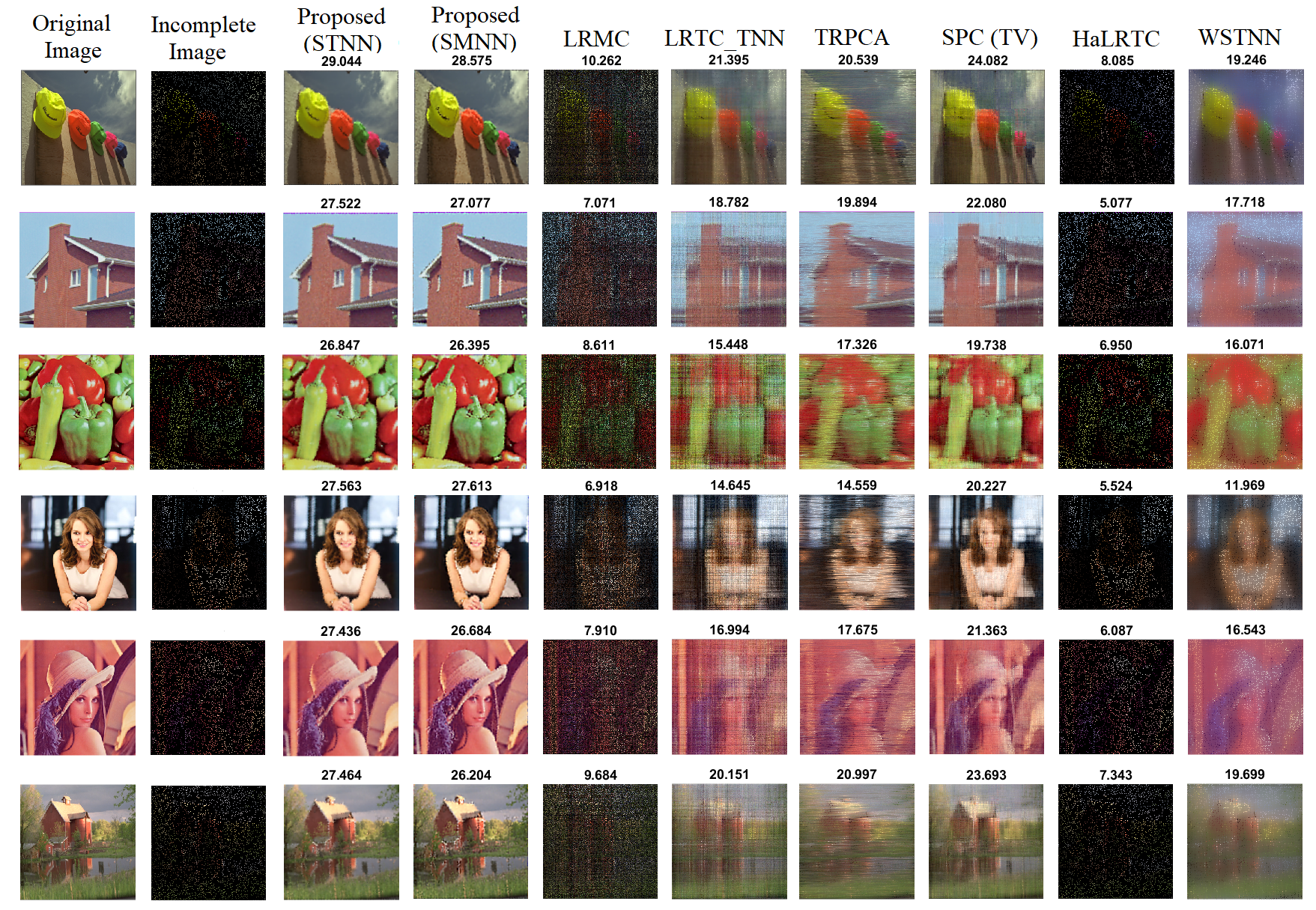} 
    \caption{PSNR comparison of different low rank completion methods with sampling 10\% of pixels for Example \ref{exa_2}.}
    \label{fig:90centroids}
\end{figure*}

\begin{figure*}[ht!]
    \centering
    \includegraphics[width=1.0\linewidth]{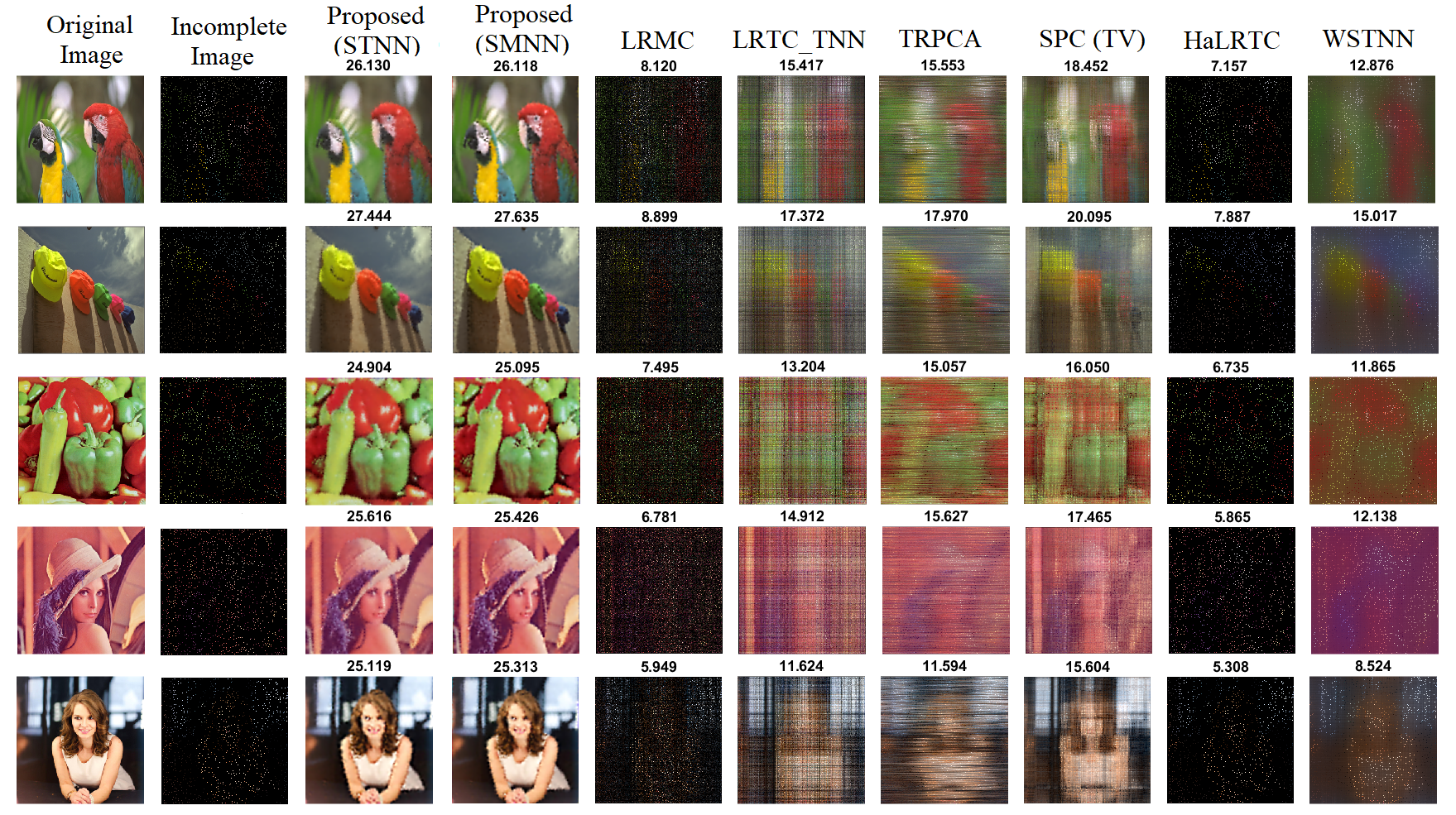} 
    \caption{PSNR comparison of different low rank completion methods with  sampling 5\% of pixels for Example \ref{exa_2}.}
    \label{fig:95centroids}
\end{figure*}

\begin{figure*}[ht!]
    \centering
    \includegraphics[width=.7\linewidth]{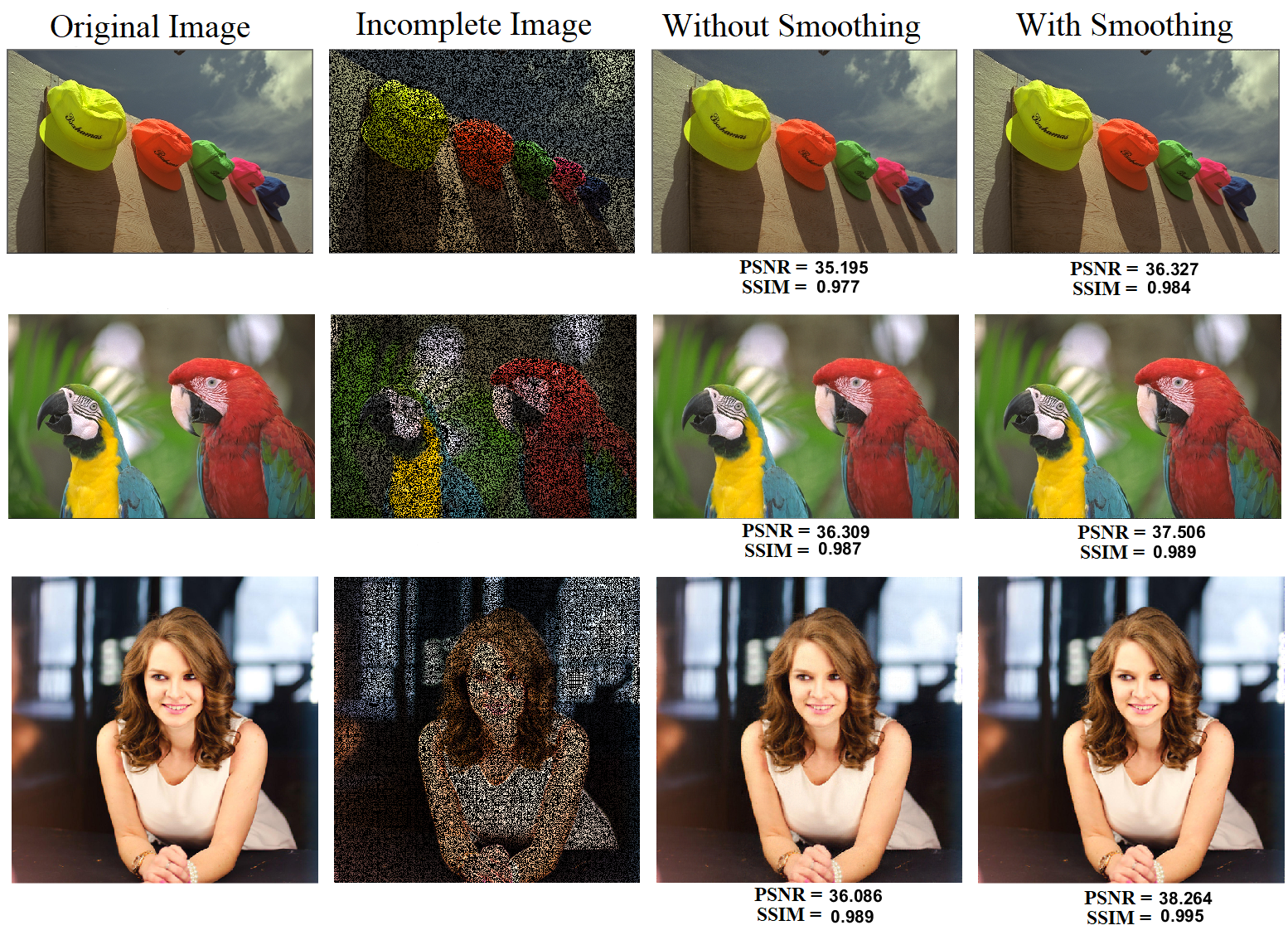} 
    \caption{Reconstruction comparison for the TNN algorithm with/without smoothing. We have  sampled $30\%$ of pixels for Example \ref{exa_2}.}
    \label{fig:compare_smooth70}
\end{figure*}

\begin{figure*}[ht!]
    \centering
    \includegraphics[width=1.0\linewidth]{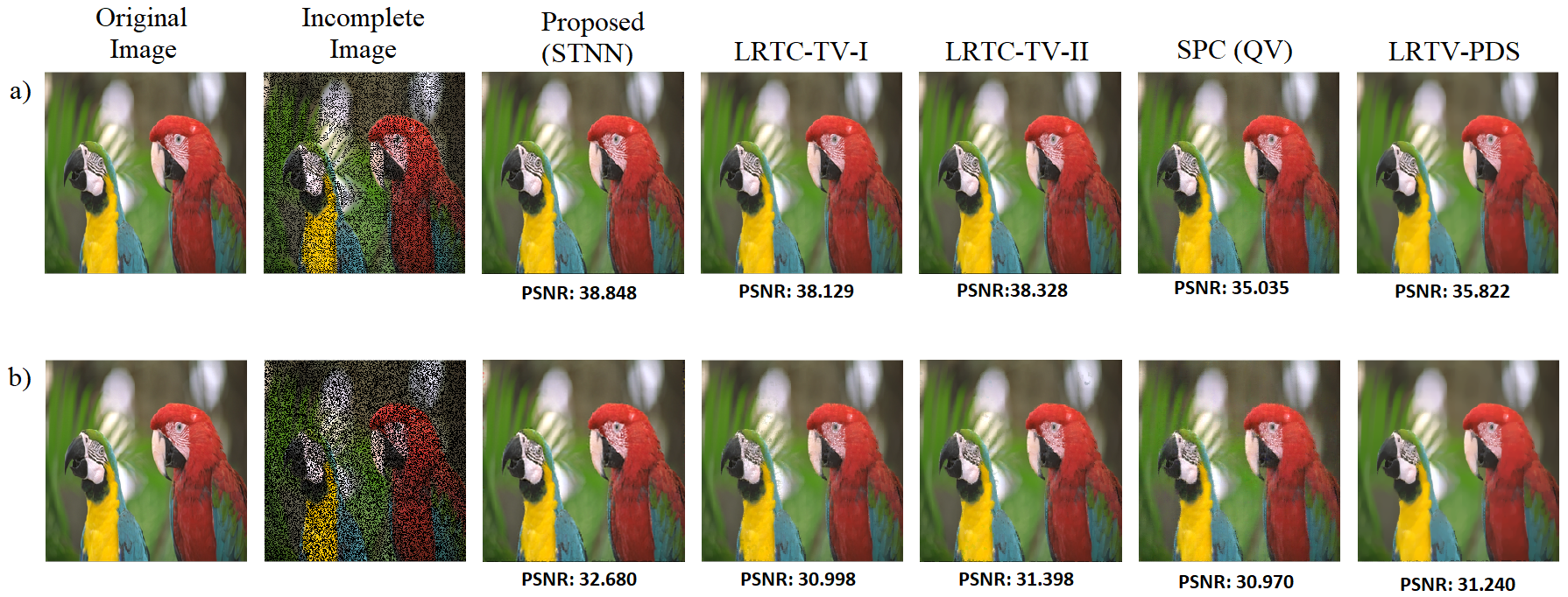} 
    \caption{Comparison of  smoothing completion algorithms with sampling 50\% of pixels for Example \ref{exa_2}. a) Centroid sampling b) Random sampling.}
    \label{fig:compare_lowranksmoothalgs50}
\end{figure*}

\begin{figure*}[ht!]
    \centering
    \includegraphics[width=1.0\linewidth]{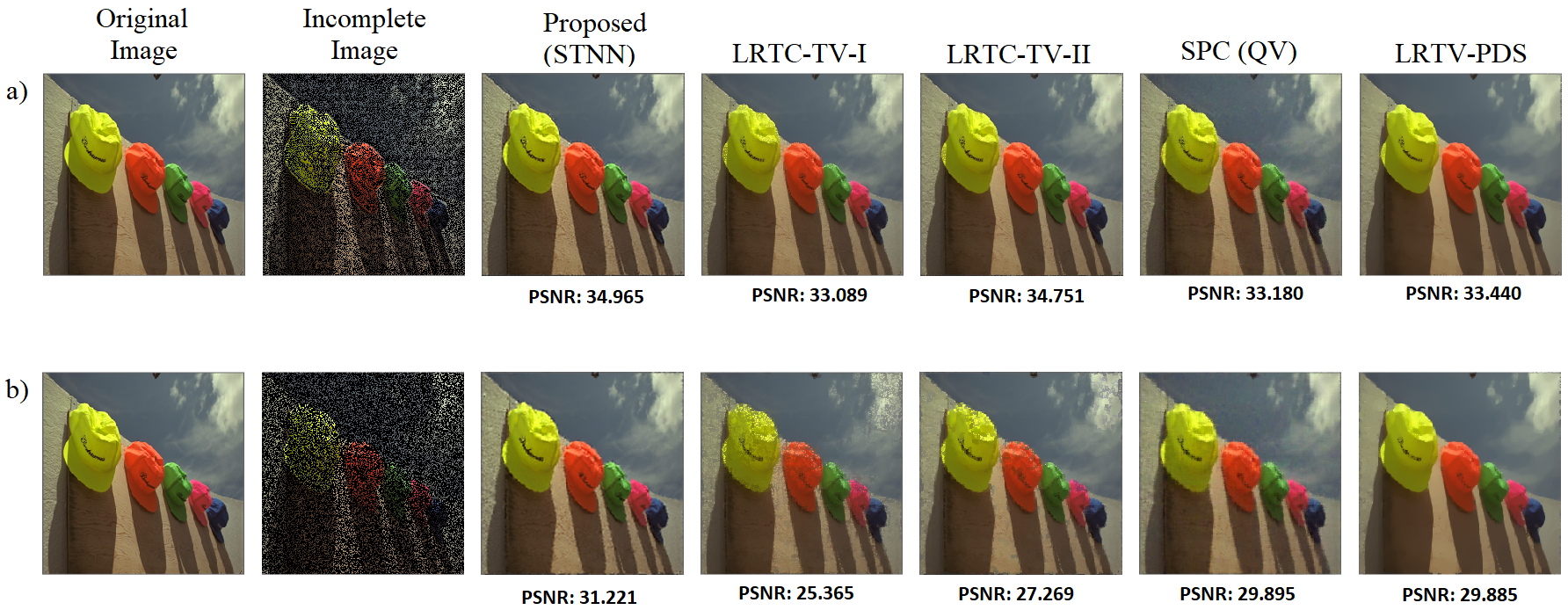} 
    \caption{Comparison of  smoothing completion algorithms with sampling 30\% of pixels for Example \ref{exa_2}. a) Centroid sampling b) Random sampling }
    \label{fig:compare_lowranksmoothalgs70}
\end{figure*}

\begin{figure*}[ht!]
    \centering
    \includegraphics[width=1.0\linewidth]{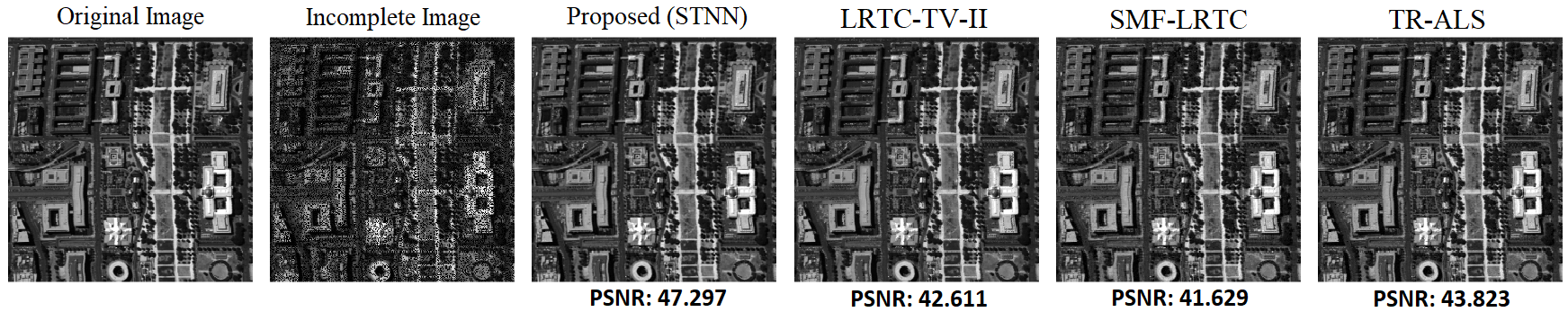} 
    \caption{PSNR comparison of  tensor completion algorithms with sampling 50\%  of pixels from the WDC hyper-spectral Data for Example \ref{exa_2}. The superpixel with centriod sampling was used.}
    \label{fig:compare_lowranksmoothalgsWDC}
\end{figure*}

\begin{exa}\label{exa_2}
{\bf (Comparing the SMNN, the STNN with the baseline completion algorithms)} 
Our main goal in this experiment is to show the better recovery performance of the proposed SMNN and STNN than the baseline completion algorithms: LRMC \cite{lu2018unified}, LRTC\_TNN \cite{lu2018unified}, TRPCA \cite{lu2019tensor},  HaLTRC\cite{liu2012tensor}, SPC(TV) \cite{yokota2016smooth} and WSTNN \cite{zheng2020tensor} which to the best of our knowledge have provided the sate-of-the-art results. All the hyper-parameters were tuned as used in the methods for fair comparison. This is to ensure that all the methods performed as best as possible. Note that since the first experiment confirmed the better performance of the centroid superpixel approach provides better results than other sampling methods, throughout this experiment we use it to sample pixels. The results PSNR of the reconstructed images for 70\%,30\%,20\%,10\% and 5\% pixel sampling are displayed in Figures \ref{fig:30centroids}-\ref{fig:95centroids}. We see that our proposed algorithms perform better than the other algorithms in most cases and for various sampling ratios. Also, the results shows that the STNN performed better than the SMNN. Moreover, to highlight the effect of filtering/smoothing scheme, we performed a new experiment with 30\%  of available pixels and presents results for three images (kodim03, kodim23 and Kate). The reconstructed images using the STNN with/without filtering are displayed in Figure \ref{fig:compare_smooth70} and the PNSR and SSIM results show that the smoothing/filtering technique can improve the recovery results.

Owing to the fact that our method uses smoothness, we also performed experiments comparing our methods with other tensor completion algorithms that incorporate smoothness.
Figures \ref{fig:compare_lowranksmoothalgs50} and  \ref{fig:compare_lowranksmoothalgs70} show results comparing  some smoothed tensor completion methods such as LRTC-TV-I \cite{li2017low}, LRTC-TV-II \cite{li2017low}, SPC(QV) \cite{yokota2016smooth}, and LRTV-PDS \cite{yokota2017simultaneous}.
The results show that our methods can provide comparable results to the mentioned smoothed tensor completion methods and even in some cases, they achieve better performance. We performed experiments on images with 50\% and 30\% of pixel sampled both using random and superpixels sampling methods. The obtained results are shown in Figures \ref{fig:compare_lowranksmoothalgs50} and \ref{fig:compare_lowranksmoothalgs70}. We see that the proposed technique can achieve better results than the baseline techniques. We can also observe that reconstruction of incomplete image whose pixels were sampled by the proposed superpixel method have better quality. To further examine the proposed smoothed completion algorithm, we used the Washington DC mall hyperspectral data \footnote{https://engineering.purdue.edu/~biehl/MultiSpec/hyperspectral.html} which is of size $1208 \times 307 \times 191$. We used a sub-tensor of $256 \times 256 \times 30$ and sampled only 50\% of its pixels. We applied the proposed superpixel sampling method with the centroid pixel selection to the first frontal slice to compute the mask operator and this mask was used for all frontal slices. Then our proposed smoothed tensor completion method, SMF-LRTC \cite{zheng2019low}, LRTC-TV-II \cite{li2017low}, and TR-ALS \cite{wang2017efficient} algorithms were applied to it to reconstruct the hyperspectral data.
Figure \ref{fig:compare_lowranksmoothalgsWDC} represents the reconstructed images obtained by the algorithms for band 20. Clearly, we see that the proposed smoothed tensor completion method provided better recovery results. 
\end{exa}

\begin{figure*}[ht!]
    \centering
    \includegraphics[width=1.0\linewidth]{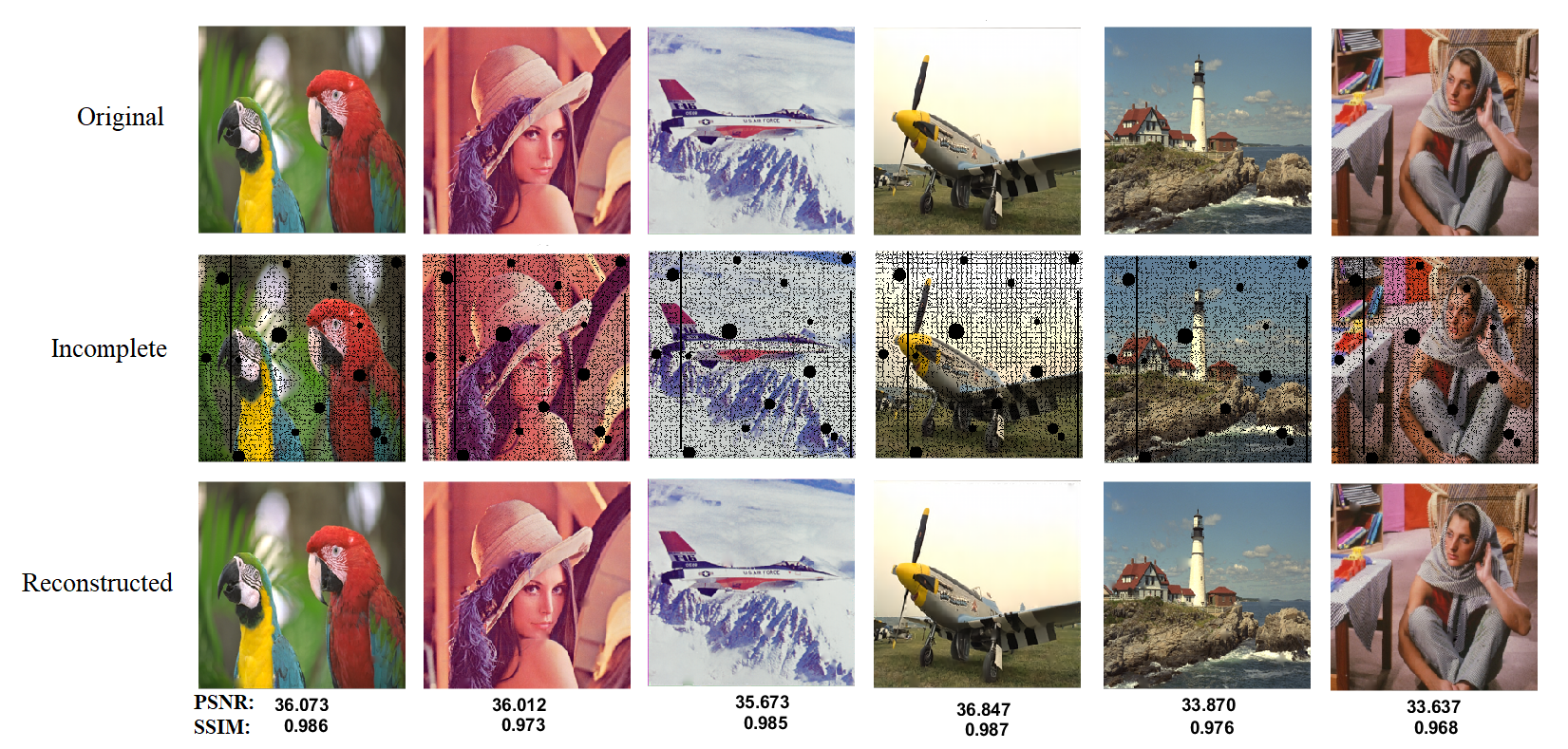} 
    \caption{The structured results of the STNN algorithm for structured missing pixels for Example \ref{exa_3}.}
    \label{fig:circles}
\end{figure*}

\begin{figure*}[ht!]
    \centering
    \includegraphics[width=1.0\linewidth]{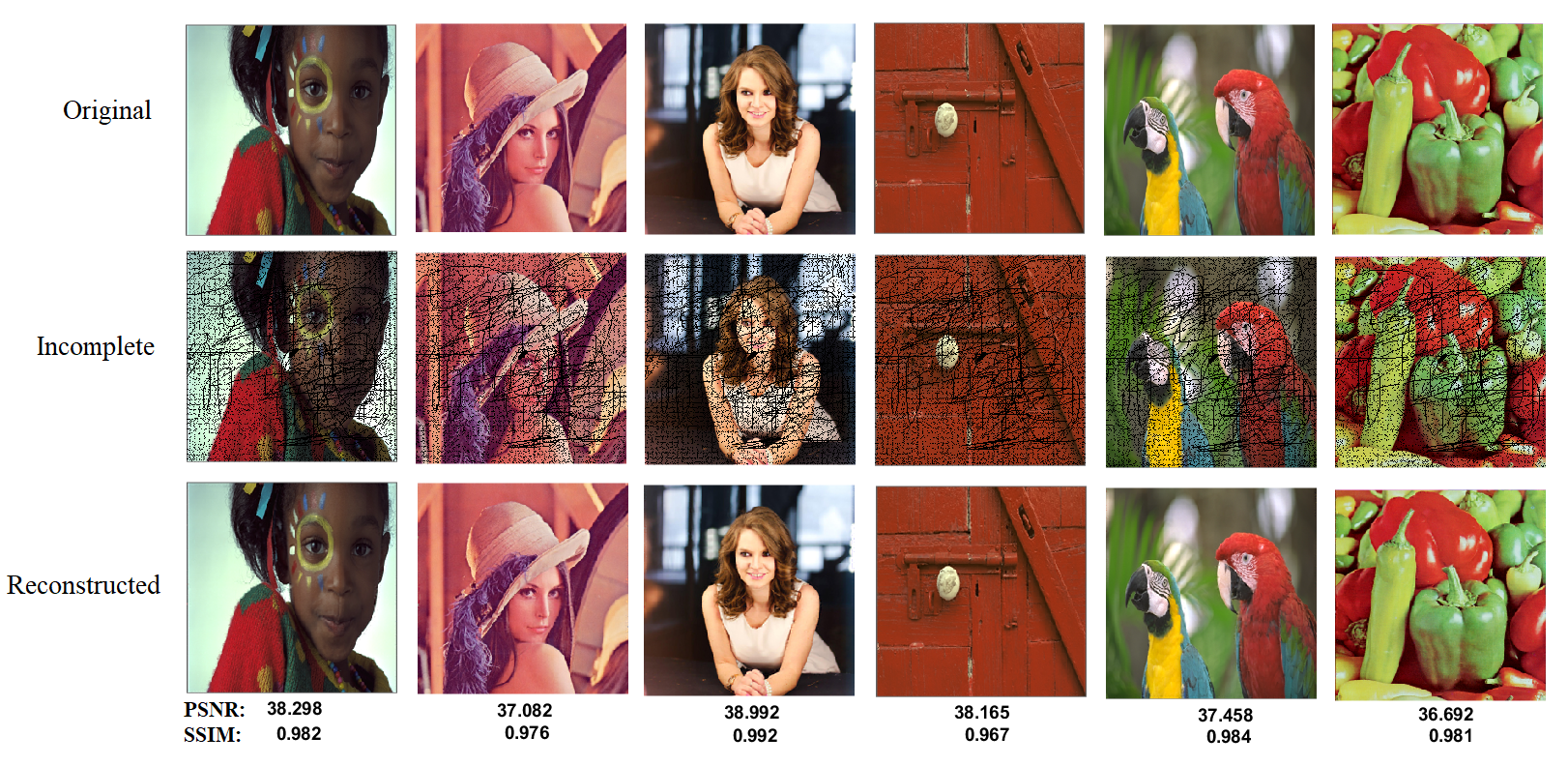} 
    \caption{The structured results of the STNN algorithm for structured missing pixels for Example \ref{exa_3}.}
    \label{fig:scratches}
\end{figure*}

\begin{figure*}[ht!]
    \centering
    \includegraphics[width=1.0\linewidth]{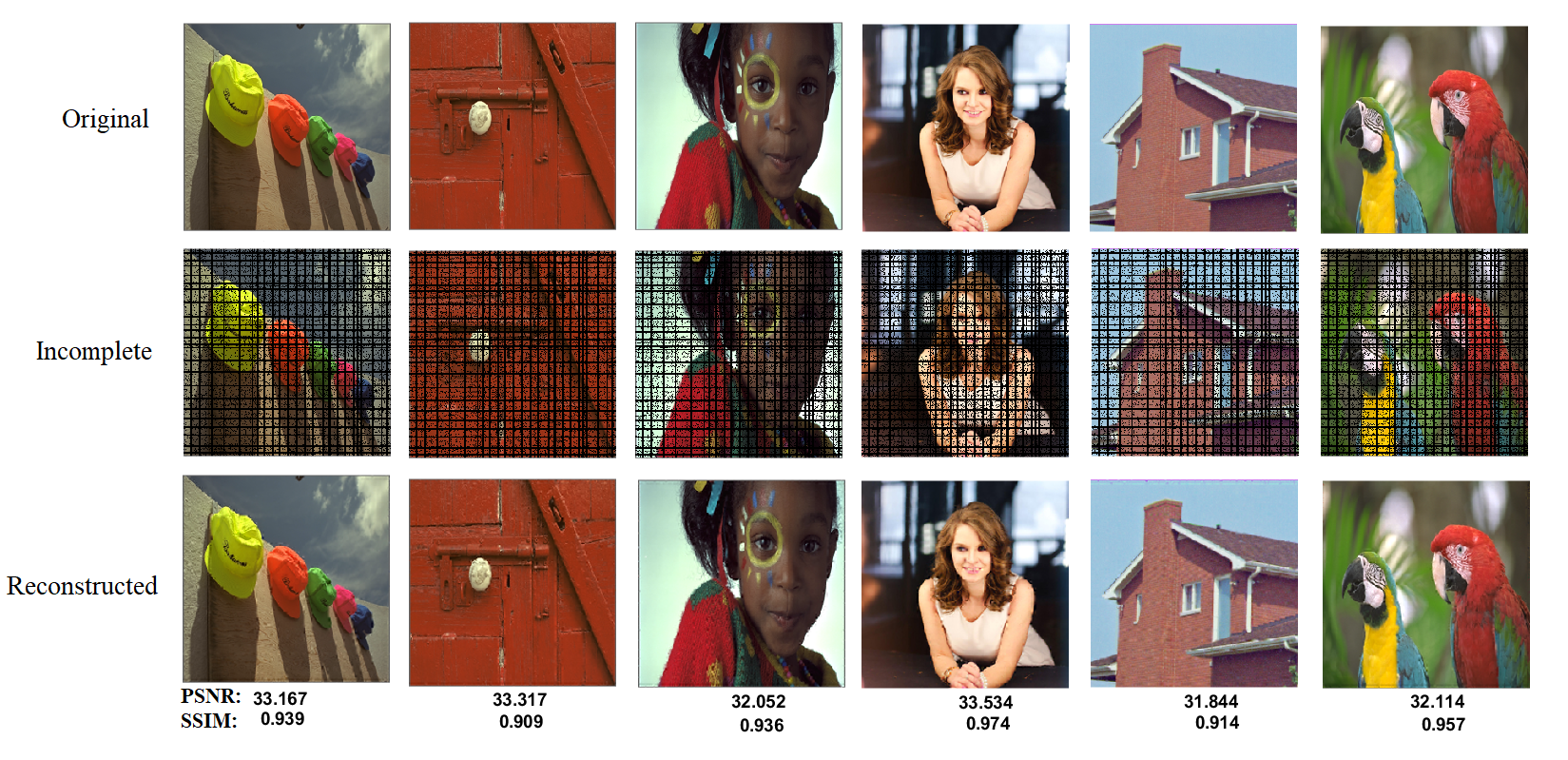} 
    \caption{The structured results of the STNN algorithm for structured missing pixels for Example \ref{exa_3}.}
    \label{fig:linesmore}
\end{figure*}

\begin{exa}\label{exa_3}
({\bf Image recovery performance of the STNN for images with structured missing pixels})
%For this experiments all the images had $30\%$ of the pixels removed using superpixel clustering and centroid selection. 
In this experiment, we evaluate the efficiency of the proposed STNN which achieved the best results among other completion algorithms for recovering images with structured missing pixels which is a more challenging case. To this end, we considered three types of structured missing patterns depicted in the first rows of Figures \ref{fig:circles}-\ref{fig:linesmore}. Then, apply STNN algorithm to estimate the missing pixels. The reconstructed images using the STNN algorithm are shown in the third rows of Figures \ref{fig:circles}-\ref{fig:linesmore}.  Clearly, the results confirmed that the STNN algorithm is also applicable for recovering images with structured missing pixels.
\end{exa}
%For this experiments, the images are distorted with different structural missing pixels. For instance, in Figure \ref{fig:circles}, $20\%$ of pixels were randomly removed in a form of black big spots and forming some holes in the images. In Figure \ref{fig:scratches} continuous scratching the images was performed to removed pixels and distort the images. Finally in Figure \ref{fig:linesmore} image, we removed more than $40\%$ of whole columns and rows of pixels randomly. The best results from these experiments are reported for both SMNN and STNN algorithms.

%Finally, we applied our proposed algorithm to an MRI images. Here, we mainly exploited four different kinds of structural missing patterns (see Figure \ref{fig:mri}).
%In the first experiment, we scratched the data image (first image in Figure \ref{fig:mri}), in the second one we removed $10\%$ of pixels and $10\%$ of columns and rows randomly (second image in Figure \ref{fig:mri}), in the third experiment, we removed $40\%$ of pixels and putting some spots in the image (third image in Figure \ref{fig:mri}), and in the fourth experiment, we removed $40\%$ columns and rows (fourth image in Figure \ref{fig:mri}). 

%Moreover, we compared the performance of our algorithm for the MRI image with some other tensor completion algorithm. These results are reported in Figure \ref{fig:mri2}. Here again, our algorithms provided better performance.

%as explained in

%\begin{figure*}[ht!]
%    \centering
%    \includegraphics[width=0.8\linewidth]{99_centroids_8_crop_tnn.png} 
%    \caption{PSNR comparison of different low rank completion methods with 99\% of pixels removed.}
%    \label{fig:99centroids}
%\end{figure*}

% \begin{figure*}[ht!]
%     \centering
%     \includegraphics[width=0.6\linewidth]{cpu_dimensions1.png} 
%     \caption{Comparison of running time using different data dimensions}
%     \label{fig:cpu_dimensions}
% \end{figure*}

% \begin{figure*}[ht!]
%     \centering
%     \includegraphics[width=0.5\linewidth]{mri_2_Resultsnew_all_50.png} 
%     \caption{PSNR comparison of different missing ratios using the STNN algorithm on MRI images}
%     \label{fig:mri_comparison}
% \end{figure*}

%\begin{figure*}[ht!]
%    \centering
%    \includegraphics[width=1.0\linewidth]{Compare.png}
%    \caption{Visual and PSNR comparison of reconstructed images using different tensor completion algorithms. In most cases, our algorithm provides best performance.}
%  \label{fig:Image Results}
%\end{figure*}
\section{Conclusion and future works}\label{Conclude:Sec}
In this work, we investigated the effects of superpixel clustering and pixel selection for the task of image completion. More precisely, we proposed to apply superpixel clustering method as an efficient segmentation/clustering approach to capture important textures underlying a given image in some partitions. Then we select pixels from each cluster based on different strategies, e.g., center, boundary, etc. The experiment results showed that the best results can be achieved by selecting the centroid pixel (pixel located in the center). We also formulated the tensor completion based on the tubal tensor nuclear norm and also matrix nuclear norm applied on the unfolding matrices. We equipped the algorithm with a smoothing technique to achieve better results. Extensive simulation results on a variety of images show the effectiveness and applicability of the proposed algorithm. 
%Experiments conducted proved the applicability of the method.
In the future work, we will use the truncated tubal nuclear norm \cite{xue2018low} in minimization problem \eqref{rank_minimization_tensor}. Also acceleration of the proposed algorithm using the randomization technique is an interesting topic needs to be investigated. 
%Consider and image$\bf X$ that has been clustered into $K$ super-pixel clusters with $C$ centroids where $k=1,\ldots, K$
%ie. $SP_1, \ldots,SP_K$ and $c_1=(z_1,y_i), \ldots,c_K=(z_k,y_k)$
%For each cluster $SP_k$, we find the best centroid by:
%\begin{equation}
%\begin{array}{l}
%\mathop {\min }
%     loss \,(sampling (\left \| C_i - C_k \right\|^2),\\
%    s.t \,\, C_i \,\,\,in \,\, SP_k \\
%    and \\
%    \left \|  << C_1 \cdots C_k >> - {\bf X}  \right\|_F < %\epsilon.
%\end{array}    
%\end{equation}

%where $gen<< .>>$ transforms the clusters to form an image$\hat{\bf X}$ and $\epsilon$ is the threshold
% \ifCLASSOPTIONcaptionsoff
%   \newpage
% \fi
\clearpage
\bibliographystyle{IEEEtran}

\bibliography{IEEEabrv,biblio.bib}

\section*{Conflict of Interest Statement
}The authors have no conflicts of interest to declare. 
\clearpage
 \appendix \label{Appendix:Sec}
 \section{Basic Definitions and Algorithms}
We provide definitions to some concepts used in our paper.
We also provide some algorithms we adopted for our paper.
\begin{defn} ({t-product}) 
The t-product of two tensors  $\underline{\mathbf X}\in\mathbb{R}^{I_1\times I_2\times I_3}$ and $\underline{\mathbf Y}\in\mathbb{R}^{I_2\times I_4\times I_3}$, is given by $\underline{\mathbf C}\in\mathbb{R}^{I_1\times I_4\times I_3}$, this is defined as
\begin{equation}\label{TPROD}
\underline{\mathbf C} = \underline{\mathbf X} * \underline{\mathbf Y} = {\rm fold}\left( {{\rm circ}\left( \underline{\mathbf X} \right) \, . \, {\rm unfold}\left( \underline{\mathbf Y} \right)} \right),
\end{equation}
where 
\[
{\rm circ} \left(\underline{\mathbf X}\right)
=
\begin{bmatrix}
\underline{\mathbf X}(:,:,1) & \underline{\mathbf X}(:,:,I_3) & \cdots & \underline{\mathbf X}(:,:,2)\\
\underline{\mathbf X}(:,:,2) & \underline{\mathbf X}(:,:,1) & \cdots & \underline{\mathbf X}(:,:,3)\\
 \vdots & \vdots & \ddots &  \vdots \\
 \underline{\mathbf X}(:,:,I_3) & \underline{\mathbf X}(:,:,I_3-1) & \cdots & \underline{\mathbf X}(:,:,1)
\end{bmatrix},
\]
and 
\[
{\rm unfold}(\underline{\mathbf Y})=
\begin{bmatrix}
\underline{\mathbf Y}(:,:,1)\\
\underline{\mathbf Y}(:,:,2)\\
\vdots\\
\underline{\mathbf Y}(:,:,I_3)
\end{bmatrix},\hspace*{.5cm}
\underline{\mathbf Y}={\rm fold} \left({\rm unfold}\left(\underline{\mathbf Y}\right)\right).
\]
\end{defn}

%$·$  is the matrix product.
It can be seen that the t-product operation \eqref{TPROD} is equivalent to the circular convolution operator, and can therefore be easily computed through  the Fast Fourier Transform (FFT). To be precise, all tubes from the two tensors $\underline{\bf X},\,\underline{\bf Y}$ are transformed into the frequency domain, then the  frontal slices of the spectral tensors are multiplied. we then apply the Inverse Fast Fourier Transform (IFFT) to all the tubes in the last tensor. The t-product can also be written in the Fourier domain as follows:
\begin{equation}\label{TPROD_fourier}
\underline{\mathbf C} = \underline{\mathbf X} * \underline{\mathbf Y} \Leftarrow\Rightarrow \widehat {\mathbf C} = { \widehat{\mathbf X} \widehat {\mathbf Y} } ,
\end{equation}

where $\widehat{\mathbf X}, \widehat {\mathbf Y}$ and $\widehat {\mathbf C}$ are block diagonal matrices defined as follows: 
\[
\widehat{\mathbf X} = bdiag(\underline{\widehat{\mathbf X}})=
\begin{bmatrix}
\widehat{\mathbf X}^{(1)} &  &  & \\
 & \widehat{\mathbf X}^{(2)} &   & \\
 &  & \ddots &   \\
 &  &  & \widehat{\mathbf X}^{(I_3)}
\end{bmatrix}
\]
where $\widehat{\mathbf X}^{(1)}$ is the a matrix computed by applying the fast Fourier transform. The operator $bdiag(.)$ maps the tensor $\underline{\widehat{\mathbf X}}$ to the block diagonal matrix $\widehat{\mathbf X}$. The procedure  for t-product in the Fourier domain is summarized in Algorithm \ref{ALG:TSVDP}. 

\RestyleAlgo{ruled}
\LinesNumbered
\begin{algorithm}
\SetKwInOut{Input}{Input}
\SetKwInOut{Output}{Output}\Input{Two data tensors $\underline{\mathbf X} \in {\mathbb{R}^{{I_1} \times {I_2} \times {I_3}}},\,\,\underline{\mathbf Y} \in {\mathbb{R}^{{I_2} \times {I_4} \times {I_3}}}$} 
\Output{t-product $\underline{\mathbf C} = \underline{\mathbf X} * \underline{\mathbf Y}\in\mathbb{R}^{I_1\times I_4\times I_3}$}
\caption{The t-product tensor in the Fourier domain \cite{kilmer2011factorization}}\label{ALG:TSVDP}
      {
      $\widehat{\underline{\mathbf X}} = {\rm fft}\left( {\underline{\mathbf X},[],3} \right)$\\
      $\widehat{\underline{\mathbf Y}} = {\rm fft}\left( {\underline{\mathbf Y},[],3} \right)$\\
\For{$i=1,2,\ldots,I_3$}
{                        
$\widehat{\underline{\mathbf C}}\left( {:,:,i} \right) = \widehat{\underline{\mathbf X}}\left( {:,:,i} \right)\,\widehat{\underline{\mathbf Y}}\left( {:,:,i} \right)$\\
}
$\underline{\mathbf C} = {\rm ifft}\left( {\widehat{\underline{\mathbf C}},[],3} \right)$   
       	}       	
\end{algorithm}

\begin{defn} ({Transpose})
The transpose of a tensor $\underline{\mathbf X} \in\mathbb{R}^{I_1\times I_2\times I_3}$ is denoted by $\underline{\mathbf X}^{T}\in\mathbb{R}^{I_2\times I_1\times I_3}$. It is obtained by applying transpose to all the frontal slices and then reversing the order of the transposed frontal slices from the second through to the last frontal slice.
\end{defn}

\begin{defn} ({Identity tensor})
An identity tensor $\underline{\mathbf I}\in\mathbb{R}^{I_1\times I_1\times I_3}$ is a tensor with first frontal slice being an identity matrix of size $I_1\times I_1$, and all other frontal slices being equal to zero.
\end{defn}

\begin{defn} ({Orthogonal tensor})
A tensor $\underline{\mathbf X}\in\mathbb{R}^{I_1\times I_1\times I_3}$ is orthogonal if ${\underline{\mathbf X}^T} * \underline{\mathbf X} = \underline{\mathbf X} * {\underline{\mathbf X}^ T} = \underline{\mathbf I}$ is satisfied.
\end{defn}

\begin{defn} ({f-diagonal tensor})
An f-diagonal tensor is a tensor with all of its frontal slices being diagonal.
%If each of a tensor's frontal slices is a diagonal matrix, it is called f- diagonal. 
\end{defn}

\begin{defn} ({t-SVD})
A tensor $\underline{\mathbf X}\in\mathbb{R}^{I_1\times I_2\times I_3}$, can be decomposed as
\[
\underline{\mathbf X} = \underline{\mathbf U} * \underline{\mathbf S}  * {\underline{\mathbf V}^T},
\]
where $\underline{\mathbf U}\in\mathbb{R}^{I_1\times I_1\times I_3}$, $\underline{\mathbf V}\in\mathbb{R}^{I_2\times I_2\times I_3}$ are orthogonal tensors, and tensor $\underline{\mathbf S}\in\mathbb{R}^{I_1\times I_2\times I_3}$ is f-diagonal. %Refer to \ref{fig:t_SVD_illustration} for illustration.
\end{defn}
%In the next section, we briefly introduce the matrix column selection and  approaches. They are used in our subsequent cross tensor-analysis.
%It is seen the main expensive part of the algorithm is computation of truncated SVD and later we discuss two randomized approach to reduce the computational complexity of the deterministic Algorithm \ref{ALG:t-SVD}.
\RestyleAlgo{ruled}
\LinesNumbered
\begin{algorithm}
\SetKwInOut{Input}{Input}
\SetKwInOut{Output}{Output}\Input{A data tensor $\underline{\mathbf X} \in {\mathbb{R}^{{I_1} \times {I_2} \times {I_3}}}$ and target tubal rank $R$} 
\Output{${\underline{\mathbf U}_R} \in {\mathbb{R}^{{I_1} \times R \times {I_3}}},\,\,{\underline{\mathbf S}_R} \in {\mathbb{R}^{R \times R \times {I_3}}},\,\,{\underline{\mathbf V}_R} \in {\mathbb{R}^{{I_2} \times R \times {I_3}}}$}
\caption{Truncated t-SVD \cite{kilmer2013third}}\label{ALG:t-SVD}
      {
$\widehat{\underline{\mathbf X}}= {\rm fft}\left( {\underline{\mathbf X},[],3} \right)$\\
\For{$i=1,2,\ldots,I_3$}
{                        
$[{\mathbf U},{\mathbf S},{\mathbf V}] = {\rm truncated\_ svd}\left( {\bar{\underline{\mathbf X}}(:,:,i),R} \right)$\\
${\widehat{\underline{\mathbf U}}}\left( {:,:,i} \right) = {{\mathbf U}}$\\
${\widehat{\underline{\mathbf S}}}\left( {:,:,i} \right) = {{\mathbf S}}$\\ 
${\widehat{\underline{\mathbf V}}}\left( {:,:,i} \right) = {{\mathbf V}}$\\
}
${\underline{\mathbf U}} = {\rm ifft}\left( {{\widehat{\underline{\mathbf U}}},[],3} \right),\,{\underline{\mathbf S}} = {\rm ifft}\left( {{\widehat{\underline{\mathbf S}}},[],3} \right),\,{\underline{\mathbf V}} = {\rm ifft}\left( {\widehat{\underline{\mathbf V}},[],3} \right)$  
       	}       	
\end{algorithm}

% \todo[inline]{I think we do not need Figure 1 because we do not have any tensor network}
% \begin{figure*}[ht!]
%     \centering
%     \captionsetup{justification=centering}
%     %\subfigure[Graphical representations of tensors and matrices]
%     {
%         \includegraphics[width=0.8\linewidth]{PIC1.png}
%     }
%     \hspace{20cm}\caption{Graphical representations of tensors and matrices \cite{cichocki2016low}} 
%     \label{fig:Graphrepres}
% \end{figure*}

% \begin{figure*}[ht!]
%     \centering
%     \includegraphics[width=0.8\linewidth]{for_paper_tSVD_crop.png} 
%     \caption{Graphical Illustration of the t-SVD of an $I_1 \times I_2 \times I_3$ tensor \cite{xue2018low}}
%     \label{fig:t_SVD_illustration}
% \end{figure*}

%Fig. 1 illustrates the distinction and connection between SVD of a matrix and t-SVD of a tensor.
The t-SVD can be obtained using the SVD of frontal slices of the original data tensor in the Fourier domain. The algorithm for computing the t-SVD for tensors is outlined in Algorithm \ref{ALG:t-SVD}. So, for the t-SVD of $\underline{\mathbf X}$, we have:
\begin{equation}\label{SVD_bdiag}
\widehat{\mathbf X}^{(i)}= \widehat{\mathbf U}^{(i)} *\, \widehat{\mathbf S}^{(i)} * \, (\widehat{\mathbf V}^{(i)})^T , \, \, \, i = 1, 2, \ldots, I_3,
\end{equation}
with $\widehat{\mathbf X}^{(i)}$ being the $i$-th frontal slice in the Fourier domain, i.e., $\widehat{\mathbf X}^{(i)}=\widehat{\mathbf X}(:,:,i)$.  
%of the tensor $\underline{\mathbf X}$
%From equation \eqref{SVD_bdiag}, we can introduce the definition of tensor tubal rank and tensor nuclear norm, which are convex envelopes for finding the tensor rank. 

Now, we introduce the tensor nuclear norm based on the t-product. 
\begin{defn} ({tensor tubal rank and nuclear norm}) \cite{xue2018low}
The tensor tubal rank of $\underline{\mathbf X}\in\mathbb{R}^{I_1\times I_2\times I_3}$ is defined as the maximum rank among all frontal slices of an f-diagonal tensor $\underline{\mathbf S} $, i.e., max $rank(\underline{\mathbf S}^{(i)})$.
Additionally, the tensor nuclear norm $\left\| {\underline{\bf X}} \right\|_\ast$ is defined as the sum of the
singular values in all frontal slices of $\underline{\mathbf S} $, i.e.,
\begin{equation}\label{nuclear1}
    \left\| {\underline{\bf X}} \right\|_\ast = {\rm tr}(\underline{\mathbf S}) = \sum_{i=1}^{I_3} {\rm tr}({\mathbf S}^{(i)}),
\end{equation}
where $\underline{\mathbf S}^{(i)}$ were defined in \eqref{SVD_bdiag}.
\end{defn}
It is shown in \cite{xue2018low} that the trace of tensor product $(\underline{\mathbf X} \ast \underline{\mathbf Y})$ equals to the trace of the product of $\widehat{\bf X}^{(1)}$ and $\widehat{\bf Y}^{(1)}$, that is
\begin{equation}\label{nuclear2}
{\rm tr}(\underline{\mathbf X} \ast \underline{\mathbf Y}) = {\rm tr}( \widehat{\bf X}^{(1)} \widehat{\bf Y}^{(1)} ).
\end{equation}
Then, it is proved \cite{xue2018low} that the tensor nuclear norm defined in \eqref{nuclear1} can be simplified as
%Meanwhile, Lu et al. [38] also identified the tensor nuclear norm derived from the t-product, that is
\begin{equation}
   \left\| {\underline{\bf X}} \right\|_\ast = {\rm tr}({\underline{\bf S}})  = {\rm tr}(\widehat{\bf S}^{(1)})=  \left\|  (\widehat{\bf X}^{(1)}) \right\|_\ast.
\end{equation}
% In this work, we use tensor nuclear norm to depict the low-rank property of a tensor.

\begin{defn}(Tensor singular value thresholding)
The singular value thresholding (SVT) \cite{xue2018low} operator $\underline{\bf D}_\beta(.)$ is performed on each frontal slice of the f-diagonal tensor $\widehat{\underline{\bf S}}$. That is,
\begin{equation}
    \underline{\bf D}_\beta(\underline{\bf X})  = {\underline{\bf U} \ast \underline{\bf D}_\beta(\underline{\bf S}) \ast \underline{\bf V}}^{T}
\end{equation}
where $\underline{\bf D}_\beta(\underline{\bf S})$ is the inverse FFT of $\underline{\bf D}_\beta(\widehat{\underline{\bf S}})$. and $\underline{\bf D}_\beta(\widehat{\underline{\bf S}}^{(i)})=$  $ {\rm diag}(\max\{\sigma_t -\beta,0  \}_{1\leq t\leq R}), \, i=1,\ldots, I_3$, $\beta>0$ is a constant and $R$ is the tubal rank.
\end{defn}

\RestyleAlgo{ruled}
\begin{algorithm}
\LinesNumbered
\SetKwInOut{Input}{Input}
\SetKwInOut{Output}{Output}
 %\Input{Initialize cluster centers $c_k = [l_k,a_k,b_k,x_k,y_k]^{T}$ by sampling pixels at regular grid steps $ S$}
 %\Output{Completed data tensor ${\underline{\bf X}}$}
 \caption{The SLIC method \cite{achanta2010slic}} \label{ALG:SLIC}
       \textbf{Set} clusters centers $c_k = [l_k, \, a_k, \, b_k, \, x_k, \, y_k]^{T}$ by taking regular grid steps $S$ to sample pixels \\
       \textbf{shift} Cluster centers in an $ n\times n$ neighborhood  to the lowest gradient location\\

 \While{$\epsilon \geq$ threshold }
{
       \For{each cluster center $c_k$}
       {
       \textbf{Assign} the best matching pixels from a $2S \times 2S$ square  neighborhood around the cluster center $c_k$ using the distance measure for $D_s$  in Equation \eqref{superpixeldistance}
       }
     \textbf{Compute} cluster centers \\
     \textbf{Calculate}  the residual error $\epsilon$ using the $L_1$ distance between recomputed centers and previous centers. 
}
 \end{algorithm}

\end{document}